\documentclass[10pt, a4paper]{article}
\usepackage[final]{lrec2026} 
\usepackage{amssymb}
\usepackage{amsmath}
\usepackage{booktabs}
\usepackage{xcolor}
\usepackage{longtable}
\usepackage{graphicx}
\usepackage{array}
\usepackage{colortbl}
\usepackage{placeins}
\usepackage{titlesec}

\usepackage{multirow}
\usepackage{siunitx}

\title{Not All News Is Equal: Topic- and Event-Conditional Sentiment from Finetuned LLMs for Aluminum Price Forecasting}

\name{Alvaro Paredes Amorin$^{1,2}$, Andre Python$^{2,3,4}$, Christoph Weisser$^{5}$}

\address{
$^1$ International Business School, Zhejiang University \\
$^2$ Center for Data Science, Zhejiang University \\
$^3$ Centre for Human Genetics, Nuffield Department of Medicine, Oxford University \\
$^4$ School of Medicine, Zhejiang University \\
$^5$ Bielefeld School of Business, Hochschule Bielefeld (HSBI) - University of Applied Sciences and Arts \\
alvaroparedesamorin@gmail.com \quad python.andre@gmail.com \quad christoph.weisser@hsbi.de
}

\abstract{
    By capturing the prevailing sentiment and market mood, textual data has become increasingly vital for forecasting commodity prices, particularly in metal markets. However, the effectiveness of lightweight, finetuned large language models (LLMs) in extracting predictive signals for aluminum prices—and the specific market conditions under which these signals are most informative—remains under-explored. This study generates monthly sentiment scores from English and Chinese news headlines (Reuters, Dow Jones Newswires, and China News Service) and integrates them with traditional tabular data, including base metal indices, exchange rates, inflation rates, and energy prices. We evaluate the predictive performance and economic utility of these models through long-short simulations on the Shanghai Metal Exchange from 2007 to 2024. Our results demonstrate that during periods of high volatility, Long Short-Term Memory (LSTM) models incorporating sentiment data from a finetuned Qwen3 model (Sharpe ratio 1.04) significantly outperform baseline models using tabular data alone (Sharpe ratio 0.23). Subsequent analysis elucidates the nuanced roles of news sources, topics, and event types in aluminum price forecasting
 \\ \newline  }

\begin{document}

\maketitleabstract

\section{Introduction}
\label{Intro}

Aluminum is a key non-ferrous metal with widespread applications in automotive, aerospace, construction, and electronics as a result of its lightweight, high corrosion resistance, and excellent conductivity. Aluminum production represents approximately 3.5\% of the electricity consumed worldwide and contributes to approximately 1\% of global carbon emissions, making it a highly energy-intensive and strategically important commodity~\citep{Cullen2013,Yi2024,IAI2021}. Its price dynamics is influenced by a combination of supply constraints, energy costs, geopolitical developments, and demand from emerging industries such as electric vehicles and renewable energy infrastructure~\citep{Luglio2023,Bastin2024}. These factors contribute to an increase in price volatility, which presents challenges for both market participants and industrial decision-makers.

Aluminum price forecasting typically uses statistical and machine learning methods that relies on tabular (numerical) data on historical prices within a time series modeling framework~\citep{Sverdrup2015, Esangbedo2024, Oikonomou2024}. Although methods that exclusively use tabular data can capture some patterns in price variations, they cannot capture information from textual data sources, such as news headlines and analyst reports, which can provide complementary information that reflect, e.g., investor sentiment and expectations.

We investigate whether the sentiment derived from finetuned large language models (LLMs) can improve aluminum price prediction and inform trading strategies. Specifically, we construct sentiment variables from English and Chinese news headlines, classify them with finetuned LLMs, and integrate these signals with numerical time-series data to forecast monthly aluminum prices. Our study also assesses how news sources, topics, and event types can lead to variation in the relevance of signals. By combining time-series models with finetuned LLM sentiment, we identify the conditions under which textual information provides the greatest economic value, offering both methodological insights and practical guidance for commodity market forecasting.

\section{Related Work}

\subsection{Statistical and machine learning approaches to forecast metal prices} \label{forecasting_literature}

Non-ferrous prices have been commonly forecasted using statistical models applied to time series data, such as Runge Kutta methods~\cite{Sverdrup2015} and autoregressive integrated moving average (ARIMA) models~\cite{Dooley2005,kriechbaumer2024improved}. More recently, \citet{Mysen2021} showed that tree-based algorithms, such as extreme gradient boosting (XGBoost), can outperform statistical methods in predicting aluminum prices. \citet{Oikonomou2024} showed that auto-regressive light gradient-boosting machine models can further improve the predictive performance on aluminum returns over six months. Larger models based on recurrent neural network architectures, such as long-short term memory (LSTM) models~\cite{Hochreiter1997}---a type of neural network model particularly effective at capturing long-term dependencies within time series data~\cite{dynamic_neural_networks}---have shown promising results in predicting aluminum prices ~\cite{Esangbedo2024}. Both models integrating several machine learning techniques, also-called hybrid models~\cite{li2023nonferrous}, and models aggregating the outputs of multiple models, which refer to ensemble models~\cite{Esangbedo2024}, showed promising results in the forecast of aluminum and other non-ferrous metal prices.

\subsection{Role of textual data in stock price forecast}

Understanding market sentiment can provide a valuable context for interpreting analyst forecasts \citet{Chen2020}. For example, \citet{Kumar2021, thormann_2021_lstm_twitter, kant_2024_neobroker} find that the integration of sentiment scores with traditional financial indicators can improve stock price forecasts. When combined with data on copper and aluminum commodities prices, \citet{Sinatrya2022} found that sentiment analysis can help predict the value of metal industry companies on the stock market. While \citet{Chen2021} showed that social media sentiment improves the accuracy of traditional econometric models for copper price forecasting, \citet{Gupta2020} found that sentiments derived from Twitter and news articles positively correlates with short-term prices of gold and silver. Within a hybrid model framework \citet{Yuan2020} added a module on the opinion score from a Chinese news website that improved the predictive accuracy of short-term gold prices. 

\subsection{A growing influence of large language models}

Large language models (LLMs) play an increasing role in the prediction of financial assets. Without the need for explicit rules, LLMs can capture sentiments from various languages by learning contextual representations via pre-training on massive corpora.

Several studies demonstrate that finetuned LLM-based sentiment signals outperform traditional financial NLP benchmarks when used in forecasting tasks. \citet{b2} show that an LLaMA3 model finetuned on FinancialPhraseBank leads to a superior sentiment classification performance compared to the base LLaMA3 model. \citet{b7} further report that instruction-tuned LLaMA 7B models provide more informative signals for downstream financial prediction tasks than non-instructed variants. Similarly, \citet{b4} find that combining fine-tuning with retrieval-augmented generation improves the economic relevance of model outputs. Applying this to a portfolio comprised of 417 stocks from the S\&P 500, \citet{konstantinidis2024finllama} obtained the highest returns with their finetuned LLaMA2 model, outperforming other lexicon-based methods and FinBERT.

The effectiveness of adaptation techniques is also observed across learning regimes. \citet{b9} show that fine-tuning Flan-T5 models significantly improves forecasting-related performance compared to zero- and few-shot setups, while adapted open-source LLMs such as LLaMA3, Mistral and Phi consistently outperform finance baseline models. Comparable gains from fine-tuning are reported by \citet{b8} in multiple open-source LLMs, with LLaMA2 and MPT producing the strongest downstream results.

Finetuned LLMs can efficiently capture market-relevant information not only from English but also from, e.g., Chinese textual data \citep{b3}. More recently, \citet{ParedesAmorin2025} show that DeepSeek, Qwen, and LLaMA models fine-tuned on English and Chinese languages can consistently outperform the benchmark financial model FinBERT, which is a pre-trained BERT-based language model specifically designed to analyze sentiment and extract information from financial text. 

\section{Methods}
\begin{figure*}[!ht]
\begin{center}
\includegraphics[width=\textwidth]{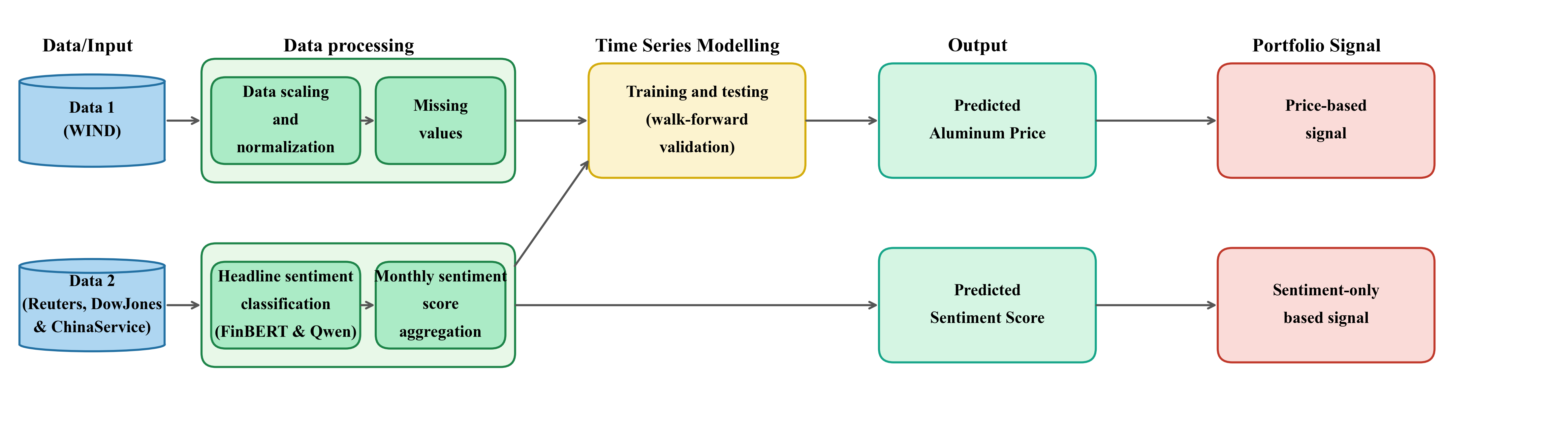}
\caption{\textbf{Workflow.} \emph{Data 1}: financial data from WIND terminal includes tabular data extracted from March 2007 to April 2024 (4,152 rows). \emph{Data 2}: textual data that includes headlines from two news sources in English (Reuters (N=4,963), Dow Jones Newswires (N=11,581), and a news source in Chinese (China News Service (N=8,970)) collected from March 2007 to April 2024. The data processing and sentiment analysis (green) includes data scaling, normalization, and the treatment of missing values for the tabular data, and the use of language models to generate new sentiment variables from \emph{Data 2}. The sentiment is classified in ``positive'', ``negative'' or ``neutral''. Monthly sentiment scores are combined with other numerical data (yellow box) to train and test time series models in order to predict monthly aluminum prices.}
\label{fig:fig1}
\end{center}
\end{figure*}

To assess the role of sentiment from news data in predicting closing aluminum prices, we compare several time series forecasting models that use tabular (``Data 1'') and/or sentiment (``Data 2'') data. The general workflow is summarized in Figure \ref{fig:fig1}.


\subsection{Data}
This research considers tabular (numerical) data gathered from the Wind terminal of the Shanghai Stock Exchange price index. It includes daily closing prices for aluminum using the Aluminum ingots commodity and factors identified as key drivers of aluminum prices~\citep{Esangbedo2024}. This includes the exchange rates of the Chinese and US currencies, the US and China inflation rates, and the closing prices of copper, zinc, and iron, as well as the prices of crude oil (OIL), Brent crude oil (LCOU5) and natural gas (NGQ5). These numerical daily data are then aggregated into monthly values during the investigated period, from March 2007 to April 2024, in line with the temporal granularity of the textual news datasets. We collected textual data related to aluminum from 2007 to 2024 from Factiva, a business information and research platform that contains a large news database. The dataset was filtered to remove noise and eliminate entries lacking relevance to aluminum price dynamics. 
The filtering method is described in Appendix \ref{filtering_sec}.
We focus on headlines (final number of headlines in parentheses)---it is common among NLP and sentiment studies to use headlines instead of full news articles~\citep{Ewald2024,Breitung2023}---in English, from Reuters (N=4,963) and Dow Jones Newswires (N=11,581), and one dataset in Chinese (mandarin) from the China News Service (N=8,970) datasets. The data used in this study from Factiva cannot be redistributed due to licensing restrictions.

\subsection{Models}

To classify aluminum news datasets, we investigate FinBERT and a lightweight LLM (Qwen3) finetuned by \citet{ParedesAmorin2025}\footnote{The source code and experimental pipeline are publicly available at \href{https://github.com/NLPforFinance/fine-tuning-of-lightweight-large-language-models}{Github}} with five different financial sentiment datasets: FinancialPhraseBank, Financial Question Answering, Gold News Sentiment, Twitter Sentiment and Chinese Finance Sentiment. These datasets cover a diverse range of financial text sources, including expert-annotated news sentences, financial document question-and-answer pairs, commodity related news, social media content, and Chinese language financial news. Here, the LLMs use the aluminum-related news to predict the sentiment of each news' headline. The models classify sentiment in three categories: ``positive'' = +1, ``negative'' = -1 \& ``neutral'' = 0. When using the finetuned Qwen model, we use the same prompt as in the finetuning process with financial sentiment datasets, and it returns one of the three labels. In the case of FinBERT, it is a BERT classifier model trained on financial data and it outputs the most likely label among the three categories by default. We computed a weighted sum of each score associated with a news item in a given month to account for the presence of multiple news items per month. For example, a score of 0.4 is given for 6 positive, 2 neutral and 2 negative news  $(6 \cdot 1 + 2 \cdot 0 + (-1) \cdot 2)/10 = 0.4)$. 

\subsection{Defining trading signal by trading strategy} \label{section:trading_strategy}

To compare sentiment-only based and price-based strategies to forecast aluminum prices, we define trading signals adapted to each strategy from which we can compare the use of news sentiment scores alone compared to the use of prices predicted by time series forecasting models using numerical (tabular) data, including sentiment scores. We define $\text{Sent}_t$ as the average sentiment score of all news headlines published during period $t$. We consider a sentiment-only based trading strategy with associated trading sentiment signal $Ss_t$ as follows:
\begin{equation}
Ss_t =
\begin{cases}
+1 & \text{if } \text{Sent}_t > 0 \quad \text{(buy/long)} \\
-1 & \text{if } \text{Sent}_t < 0 \quad \text{(sell/short)} \\
0 & \text{if } \text{Sent}_t = 0 \quad \text{(neutral)}
\end{cases}
\end{equation}

Next, we implement a price-based trading strategy using the optimal time-series forecasting model. This model was selected from various configurations via the methodology detailed in Appendix \ref{sec:timeseries_training}, with performance results provided in Appendix \ref{sec:timeseries_training_results}. 
Let $P_t^{\text{true}}$ represent the actual aluminum price at time $t$, and $P_{t+1}^{\text{pred}}$ represent the predicted price for the following period. The resulting trading signal, $Sn_t$, is defined as:

\begin{equation}
Sn_t = 
\begin{cases} 
+1 & \text{if } P_{t+1}^{\text{pred}} > P_t^{\text{true}} \quad \text{(buy/long)} \\
-1 & \text{if } P_{t+1}^{\text{pred}} < P_t^{\text{true}} \quad \text{(sell/short)} \\
0 & \text{otherwise} \quad \text{(neutral)}
\end{cases}
\end{equation}

To compute portfolio performance, the generated signals $Ss_t$ and $Sn_t$ $\in \{-1, 0, +1\}$ are multiplied by the aluminum return realized for each period.

\subsection{Evaluation Metrics}

We evaluate the economic performance of the proposed trading strategies using cumulative return and the Sharpe ratio.

Let $R_t$ denote the simple monthly return at time $t$. The cumulative return over the evaluation period $[0,T]$ is computed as:
\begin{equation}
R^{\text{cum}}_{0,T} = \prod_{t=1}^{T} (1 + R_t) - 1
\end{equation}

This metric measures the total compounded growth of the strategy over the full sample period and directly reflects long-term investment performance.

To assess risk-adjusted performance, we compute the Sharpe ratio defined as:
\begin{equation}
\text{SR} = \frac{\bar{R} - R_f}{s}
\end{equation}

where $\bar{R}$ is the average monthly return, $R_f$ is the monthly risk-free rate, and $s$ is the standard deviation of monthly returns. The Sharpe ratio captures the excess return per unit of risk, allowing comparison across competing forecasting models and trading strategies.

\section{Results and discussion}

\subsection{Role of trading strategy and market volatility} \label{sec:topic_analysis}


We assess the model performance by trading strategy (tabular-only, tabular+sentiment, sentiment only (Qwen sentiment),  sentiment only (FinBERT sentiment)) and three volatility scenarios. We partition the sample into three volatility scenarios based on a 6-month rolling standard deviation of aluminum returns, annualized by multiplying by $\sqrt{12}$. We define volatility scenarios based on fixed percentages of the observed volatility range to ensure economically meaningful thresholds. Specifically, the 20\% less volatile is considered as a low-volatility range, while the 20-50\% is considered medium-volatility and higher than the 50\% is high-volatility. In our sample, annualized volatility ranges from 1.13\% to 37.47\%, yielding threshold values of 8.40\% and 19.30\% for the low-medium and medium-high boundaries, respectively. This classification produces 66 low-volatility months, 106 medium-volatility months, and 28 high-volatility months. Appendix \ref{sec:volatility_regimes_figure} shows the resulting volatility regimes distributed across time (Panel A) as well as portfolio value of the different strategies (Panel B).

\begin{figure}[!ht]
\begin{center}
\includegraphics[width=\columnwidth]{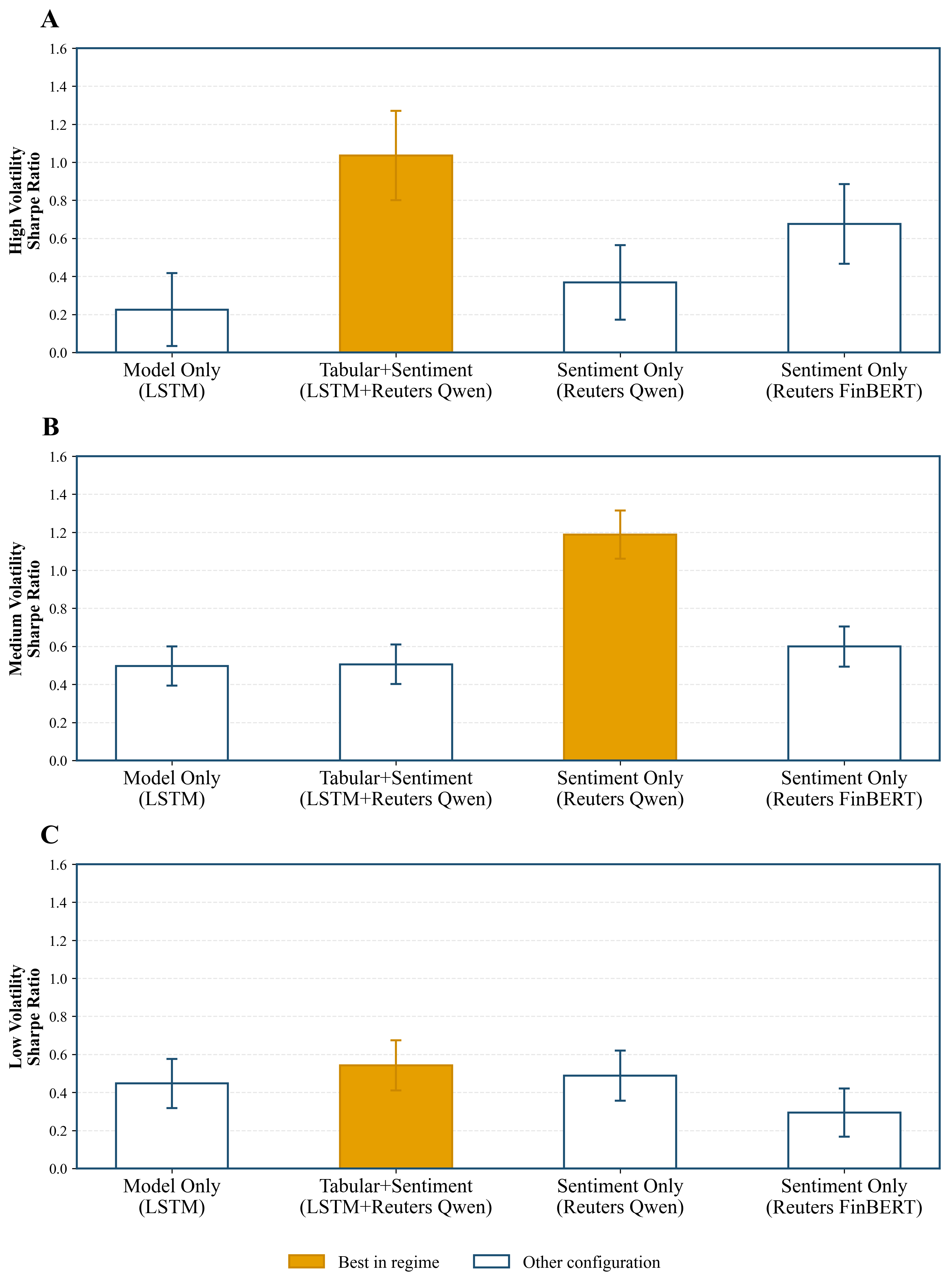}
\caption{\textbf{Evaluation of portfolio's performance by strategy and volatility scenarios.} For each strategy (tabular-only, tabular+sentiment, sentiment only (qwen), sentiment only (reuters)) the portfolio's performance is represented by the Sharpe ratio (y-axis) across three volatility scenarios (panels), with: \textbf{A} high volatility scenario (n=28 months), \textbf{B} medium volatility (n=106 months), and \textbf{C} low volatility (n=66 months). Error bars represent ±1 standard error. The highlighted bars (orange) indicate the best performing strategy within each scenario.}
\label{fig:fig31}
\end{center}
\end{figure}

Figure \ref{fig:fig31} shows the estimated Sharpe ratio for each strategy across three volatility regimes (panels A–C). During high-volatility periods, the integrated tabular and sentiment strategy achieves the highest Sharpe ratio (1.04), a 359\% improvement over the tabular-only baseline (0.23). This suggests that in turbulent markets—where historical price correlations often deviate—news sentiment captures critical directional signals, such as panic or recovery, which traditional time-series models fail to incorporate. In medium-volatility periods, the sentiment-only strategy dominates all others (Sharpe ratio = 1.19), substantially outperforming even the combined approach (0.51). This suggests that, under normal market conditions, the sentiment signal is sufficiently informative on its own and that combining it with tabular data introduces noise rather than value. In low-volatility periods, all strategies converge to similar performance levels (Sharpe ratios between 0.29 and 0.54) indicating that calm markets offer limited differentiation between approaches.

Notably, the strategy using FinBERT sentiment shows a consistent improvement as volatility increases (0.30, 0.60, 0.68), suggesting that sentiment data provide additional value with an increase in market stress. This further confirms the important role of sentiment data in predicting aluminum prices, independently of the type of models used. However, more complex models such as the finetuned Qwen model consistently outperform FinBERT, particularly during periods of medium-volatility where the performance gap is most pronounced. In summary, these findings suggest that sentiment improves the predictions of aluminum prices, with variations in its effects depending on the volatility of aluminum prices, the type of news used as a source of sentiment data, and the modeling framework. Therefore, trading strategies can benefit from weighting sentiment inputs based on textual data sources and volatility levels.

\subsection{Role of news topics and event types}
\label{subsec:topics-events}

We further investigate to what extent the topic and type of events can lead to variations in the quality of the predictive signal. We classify each Reuters headline into one of twelve topic categories (\textit{Price Movement, Environmental, Market Analysis, Production Output, Macroeconomic, Inventory Stocks, Demand Outlook, Supply Disruption, Company News, Trade Policy, Geopolitical, and Other}) with the Qwen3 8B base model through zero-shot prompts. We distinguish between predictive statements (forecasts, expectations, guidance) and statements from events that occurred in the past. To assess the contribution of individual news topics to strategy performance, we construct separate portfolios for each topic classification. Specifically, for each month, we filter headlines belonging to a given topic (e.g., \emph{Price Movement}) and compute the monthly sentiment score using only those headlines. This topic-specific sentiment score then determines the trading signal for that month. The resulting monthly returns are used to compute a Sharpe ratio for each topic, allowing us to isolate which types of news carry the most informative sentiment for aluminum price prediction. Months in which no headline of a given topic is available are excluded from that topic's portfolio, which explains the variation in sample sizes across categories.

\begin{figure}[!ht]
\begin{center}
\includegraphics[width=\columnwidth]{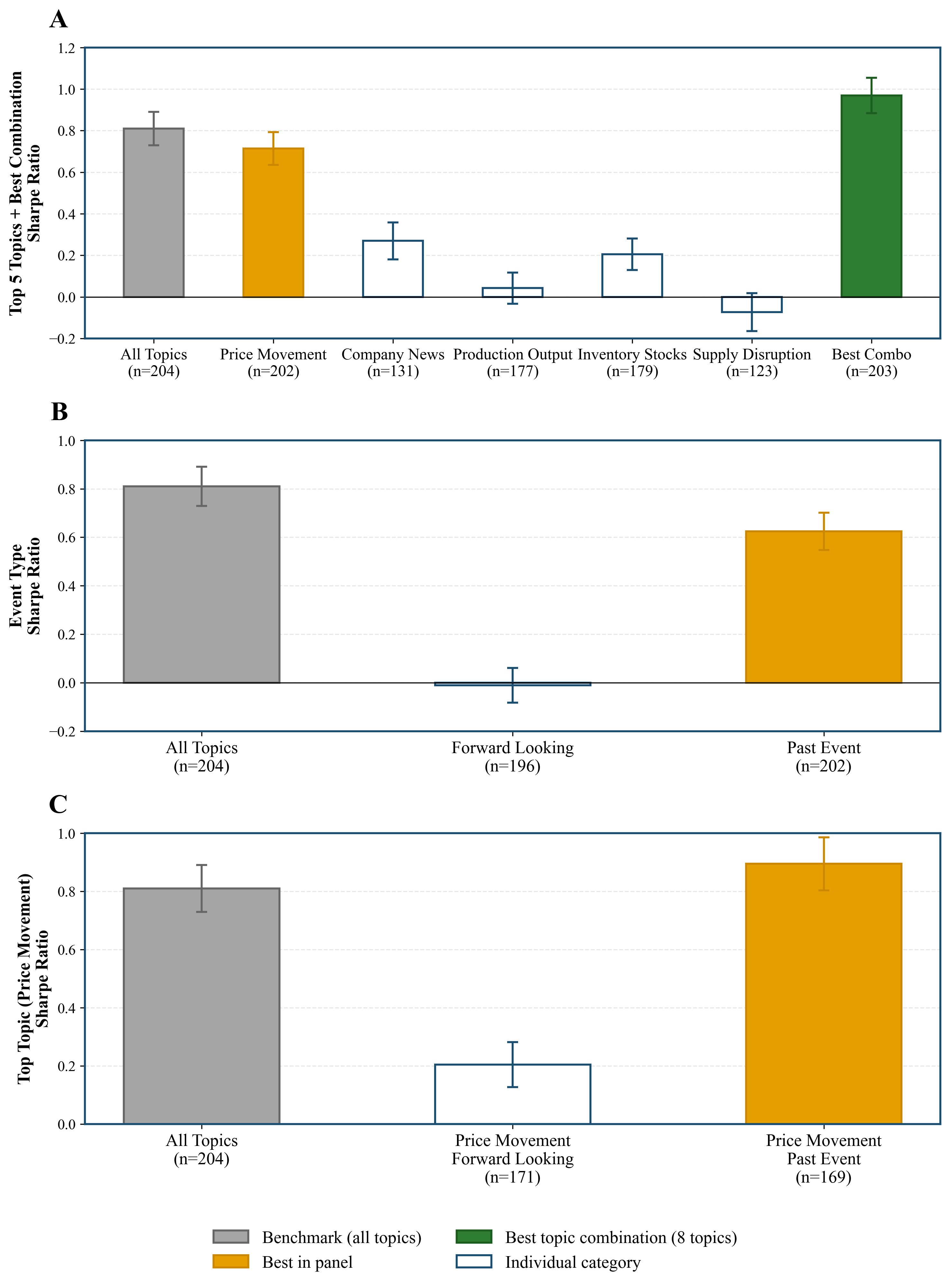}
\caption{\textbf{Evaluation of topic and event type in predictive aluminum prices.} \textbf{A} Comparison of Sharpe ratio across the top five most covered topics (Price Movement, Company News, Production Output, Inventory Stocks, Supply Disruption) and the best topic combination, all calculated using Reuters headlines with the finetuned Qwen sentiment. The benchmark (gray) represents all headlines aggregated. Here $n$ shows the number of months in which each topic is present. \textbf{B} Comparison of forward-looking versus past event news types. \textbf{C} Comparison of the top-performing topic (Price Movement) by event type. Error bars represent ±1 standard error. The green bar indicates the best-performing topic combination, while the orange bars indicate the best performing individual topic or event type.}
\label{fig:fig30}
\end{center}
\end{figure}

Figure~\ref{fig:fig30}A reports Sharpe ratios for sentiment-only strategies based on the 5 most present individual topics (\textit{Price Movement, Company News, Production Output, Inventory Stocks, Supply Disruption}) compared to the all-topics benchmark. The benchmark strategy, which aggregates sentiment across all headlines, achieves a Sharpe ratio of 0.81. In particular, no individual topic matches this benchmark performance, illustrating the diversification benefits of information aggregation. We extend this analysis by systematically evaluating all possible combinations of 2 to 11 topics to identify an optimal subset of topics, reported in the Figure as ``Best Combo''. Among 4,094 combinations tested, the best-performing subset includes eight topics: \textit{Company News}, \textit{Supply Disruption}, \textit{Inventory Stocks}, \textit{Price Movement}, \textit{Demand Outlook}, \textit{Geopolitical}, \textit{Macroeconomic}, and \textit{Other}. This combination achieves a Sharpe ratio of 1.00, representing a 23.6\% improvement over the all-topics benchmark. The four excluded topics---\textit{Production Output}, \textit{Market Analysis}, \textit{Trade Policy}, and \textit{Environmental}---appear to introduce noise that dilutes signal quality. 

Among individual topics, \textit{Price Movement} headlines generate the highest Sharpe ratio (0.71) (Figure~\ref{fig:fig30}A), which is intuitive given that such headlines directly concern price dynamics. However, the substantial gap between this topic-specific strategy and the benchmark (0.71 vs. 0.81) indicates that price commentary alone cannot account for potentially valuable signals contained in other news categories. Interestingly, \textit{Supply Disruption} headlines produce a negative Sharpe ratio ($-0.07$), suggesting that naive sentiment interpretation of disruption news can generate misleading signals, possibly because markets rapidly price in supply-side information or because the sentiment direction does not straightforwardly map to price implications.

Strategies based on forward-looking headlines achieve a nearly zero Sharpe ratio ($-0.01$), while those based on reports of events that occurred generate a Sharpe ratio of 0.62 (Figure \ref{fig:fig30}B). Forward-looking statements---analyst forecasts, company guidance, demand projections---can reflect expectations that are likely already incorporated into market prices. By contrast, reports of actual events (production figures, inventory releases, supply disruptions) can represent new information that cannot be fully anticipated. The near-zero Sharpe ratio for forward-looking content is consistent with the Efficient Market Hypothesis, suggesting that publicly available expectations are already reflected in aluminum prices \citep{fama1970efficient, tetlock2007giving}. We further decompose the dominant \textit{Price Movement} topic (Figure~\ref{fig:fig30}C). Within this category, forward-looking headlines yield a Sharpe ratio of 0.20, while occurred events achieve 0.89. This within-topic comparison reinforces the broader finding: actual outcomes carry substantially more predictive content than expectations or forecasts, even when controlling for topic.

In general, these findings suggest that sentiment strategies can benefit from filtering that emphasizes factual reporting over forward-looking commentary. This filtering could be implemented through the event-type classification framework developed in this study, allowing traders to construct signals that prioritize information content over signal volume.

\subsection{Role of the source of news}
\label{subsec:news_sources}

This section evaluates the performance of models using different news sources. Sources can vary in their coverage of topics with high predictive content, and hence lead to varying predictive performance. We compare the predictive performance of Reuters, Dow Jones and China News Service using a sentiment-only strategy with sentiment classified from aluminum related headlines by the same finetuned Qwen3 8B model. Overall, a portfolio using a sentiment-only strategy with Reuters headlines achieved a Sharpe ratio of 0.80 and 433\% Cumulative Returns across the whole time period. In contrast, portfolios based solely on sentiment extracted by the same model from headlines of Dow Jones and China News Service achieved a Sharpe ratio of 0.18 and 0.15 and Cumulative Returns of 32\% and 22\%, respectively.

\begin{figure*}[!ht]
\begin{center} 
\includegraphics[width=\textwidth]{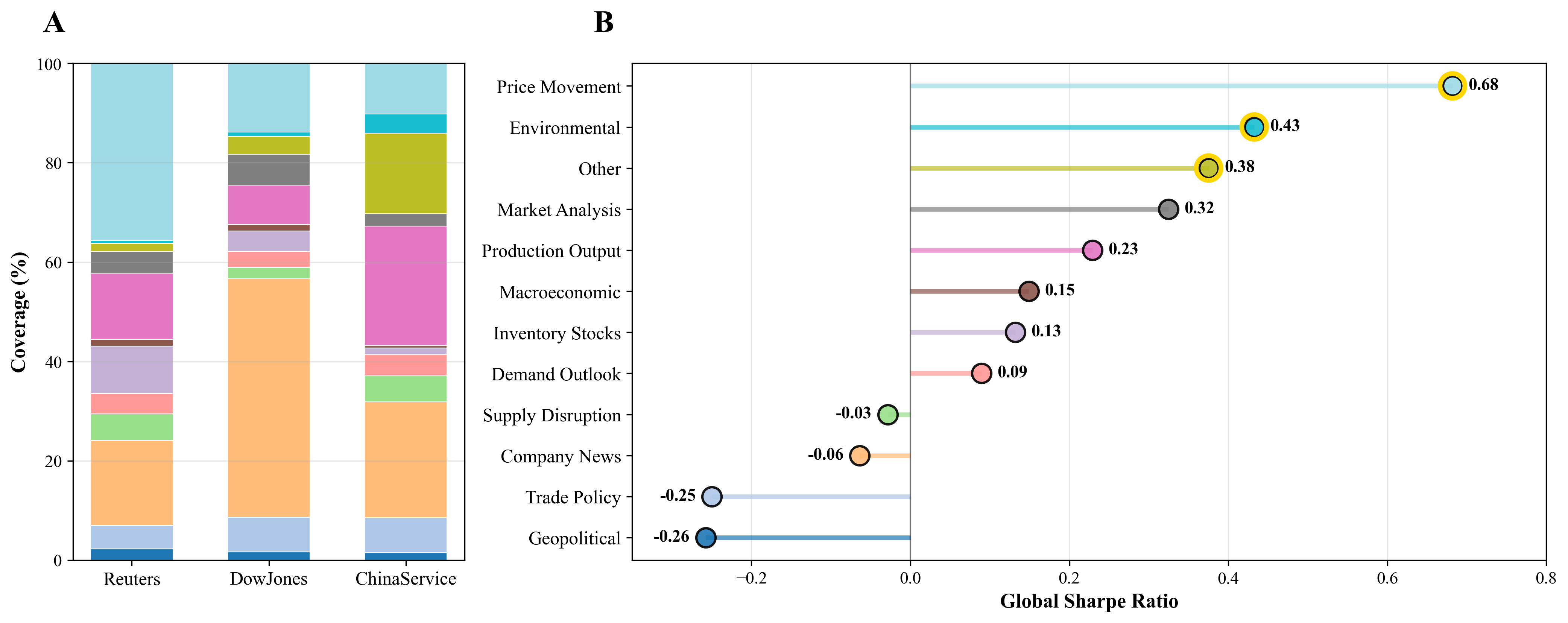} 
\caption{\textbf{Predictive performance by topic coverage. A} Percentage of headlines allocated to each topic by source. \textbf{B} Global Sharpe ratio by topic, with gold rings highlighting the top three performers. Topics are sorted by global Sharpe ratio (descending).} 
\label{fig:fig29} 
\end{center} 
\end{figure*}

Figure~\ref{fig:fig29}A reports the topic-level distribution of headlines for each source, while Figure~\ref{fig:fig29}B shows the global Sharpe ratios aggregated across all sources by topic. The results reveal that the topics with higher ability to predict aluminum price show similarity among news sources. \textit{Price Movement} seems to be the most informative category, achieving a Sharpe ratio of 0.68 and substantially outperforming all other topics, consistent with the intuition that news explicitly referencing price dynamics conveys the most direct signals for future commodity returns. \textit{Environmental} news is second (with a Sharpe ratio of 0.43), indicating that regulatory developments, climate events, and sustainability-related information contain economically meaningful signals, although it being underrepresented in overall coverage (average of 2.3\%). In contrast, \textit{Geopolitical} and \textit{Trade Policy} topics exhibit negative Sharpe ratios (-0.26 and -0.25), suggesting limited or counterproductive predictive value, potentially reflecting rapid information diffusion or anticipatory pricing effects.

The differences in overall performance between news sources closely mirror their topic coverage strategies. Reuters allocates a substantial share of its coverage to \textit{Price Movement} (35.6\%), the most predictive topic, while maintaining relatively balanced exposure in the remaining categories. By contrast, Dow Jones concentrates nearly half of its coverage (48.0\%) on \textit{Company News}, the largest single-topic allocation among all sources, despite this category showing a negative global Sharpe ratio (-0.06). Added to a limited exposure to \textit{Price Movement} (13.9\%), this results in lower predictive performance. ChinaService follows a distinct strategy, emphasizing \textit{Production Output} (24.0\%) and \textit{Company News} (23.3\%), reflecting its focus on industrial and corporate developments; while \textit{Production Output} exhibits moderate predictive ability (Sharpe ratio of 0.23), a large allocation of negatively performing \textit{Company News} reduces the overall effectiveness of sentiment-based signals.

Although allocating coverage to highly predictive topics is necessary for strong performance, it is not sufficient on its own. Table~\ref{tab:all_topics_comparison} shows the difference in the Sharpe ratio between sources on the same topics, suggesting that the quality of the information embedded in the news varies substantially between providers. For instance, within the \textit{Price Movement} category, the single most predictive topic overall, Reuters achieves a Sharpe ratio of 0.71, compared to 0.53 for DowJones and 0.67 for ChinaService. This represents a 26\% performance gap between Reuters and DowJones on identical topic exposure, highlighting that differences arise not only from topic selection, but also from how information is conveyed. Similar patterns emerge across most topics, where Reuters consistently outperforms competitors despite comparable thematic coverage. These results suggest that the advantage of Reuters might stem from higher signal-to-noise content, clearer causal structure, and more market-relevant timing, enabling sentiment signals to translate more effectively into returns. In contrast, the weaker performance of other sources on the same topics indicates dilution through descriptive, delayed, or less economically grounded reporting. 

\begin{table}[!ht]
\centering
\resizebox{\columnwidth}{!}{%
\begin{tabular}{l c c c c c}
\toprule
\textbf{Topic} & \textbf{Reuters} & \textbf{DowJones} & \textbf{ChinaService} & \multicolumn{2}{c}{\textbf{Improvement vs Reuters}} \\
\cmidrule(lr){5-6}
 & \textbf{(Benchmark)} & \textbf{Sharpe} & \textbf{Sharpe} & \textbf{DowJones (\%)} & \textbf{ChinaService (\%)} \\
\midrule

Price Movement      & 0.714  & 0.525  & 0.674  & -26.6  & -5.6  \\
Environmental       & 0.601  & 0.347  & 0.395  & -42.3  & -34.3  \\
Other               & 0.612  & 0.123  & 0.167  & -79.9  & -72.8  \\
Market Analysis     & 0.221  & 0.348  & 0.020  & +57.5 & -91.1  \\
Production Output   & 0.042  & 0.125  & 0.283  & +198.5 & +577.4 \\
Macroeconomic       & -0.051 & 0.010  & 0.164  & +120.1 & +419.8 \\
Inventory Stocks    & 0.204  & 0.027  & 0.222  & -86.9  & +8.8  \\
Demand Outlook      & 0.216  & -0.258 & 0.046  & -219.5 & -78.9  \\
Supply Disruption   & -0.073 & 0.164  & 0.124  & +323.9 & +269.6 \\
Company News        & 0.270  & -0.338 & 0.048  & -225.3 & -82.3  \\
Trade Policy        & -0.391 & -0.198 & -0.262 & +49.3 & +33.0 \\
Geopolitical        & 0.338  & -0.387 & -0.392 & -214.6 & -216.1 \\

\bottomrule
\end{tabular}
}
\caption{Sharpe ratio comparison across all individual topics, sorted by global performance.  Improvements are computed relative to Reuters as the benchmark.}
\label{tab:all_topics_comparison}
\end{table}

\section{Conclusion}
\label{sec:conclusion}

We explore the integration of sentiment signals derived from finetuned large language models (LLMs) into the prediction of aluminum prices. Although traditional time-series models that rely exclusively on tabular data provide a solid baseline for market forecasting, our results indicate that incorporating textual sentiment can enhance both predictive accuracy and economic utility, particularly under volatile market conditions. We introduce comprehensive sentiment based price forecasting of aluminum, a commodity that typically has much lower trade volume and media coverage than other metals such as gold or silver.

The finetuned Qwen3 8B consistently generates trading signals that outperform FinBERT in our out-of-sample portfolio simulations. This aligns with better sentiment classification results by finetuned LLMs compared to FinBERT reported by the literature and suggests that this is translated into improved risk-adjusted returns. Sentiment signals provided mixed value during periods of different volatility. In medium volatility regimes, signals derived from sentiment alone outperformed time series forecasting models as well as those combined with sentiment. This suggests that in certain market conditions, sentiment alone can be enough as a price directionality predictor. However, in high volatility regimes, while LSTM models using tabular data only exhibited diminished performance, when sentiment data is added, it achieves the best Sharpe ratio. These results indicate that sentiment can provide complementary information that is unlikely to be captured by numerical time-series patterns alone during volatile periods. 

This study also explores the importance of the source of sentiment. Differences in trading performance across Reuters, Dow Jones, and China News Service are explained not only by topic allocation, but also by information quality. Reuters consistently delivers higher Sharpe ratios even when controlling for topic exposure, suggesting superior signal-to-noise characteristics and more relevant framing. This finding suggests that sentiment modeling performance depends jointly on language model quality and upstream information selection. In addition, filtering news by topic can substantially improve strategy performance, as demonstrated by a 23.6\% Sharpe improvement over the all-topics benchmark when excluding certain topics.

However, we acknowledge that the sparsity of coverage of aluminum related events by news sources limits the temporal granularity of this study to monthly frequency. Future research could extend this framework to other metals or commodities as well as integrate higher frequency sentiment streams to capture finer temporal dynamics. Overall, our results provide evidence that carefully calibrated sentiment analysis using finetuned LLMs can serve as a complementary and in some cases the main source of predictive information in commodity markets, particularly under conditions where traditional numerical indicators are less informative.

\section{Acknowledgements}
This research was supported by the National Natural Science Foundation of China (Grant Nos. T2350610281 and 82273731).

\section{Bibliographical References}\label{sec:reference}

\bibliographystyle{lrec2026-natbib}
\bibliography{refs}

\clearpage

\appendix
\renewcommand{\thesection}{\Alph{section}} 

\titleformat{\section}[block]
  {\normalfont\bfseries\large}  
  {Appendix \thesection:}        
  {1em}                           
  {}


\section{Filtering aluminum news data with LLaMA} \label{filtering_sec}
News on aluminum are directly collected from Factiva from Dow Jones Newswires and Reuters. However, since it is a bulk retrieval of 7000-10000 news per source, there are many entries that do not contain any relevant information and can impact the sentiment score results. Therefore, the two aluminum news datasets are filtered using the LLM LLaMA3 8B in a few-shot scenario. 

The figure below shows the UMAP-HDBSCAN Clustering of both datasets before and after applying the filter. The clustering figure shows how after applying this filter, outliers are reduced substantially. Below each graphic, some examples from 5 different cluster groups can be seen. Sentences in the same group have high similarity, as can be seen in the examples below. In the filtered datasets, the number of clusters is reduced because some of the clusters are removed from the dataset. However, the method used with LLaMA3 does not discern between clusters but only the overall meaning of the sentence, and it is highly influenced by the examples given in the few-shot prompt. Therefore, combined with professional expertise, this method can be highly efficient in filtering large amounts of data in a fraction of time compared to doing it manually or with word mapping strategies.

\begin{figure}[!ht]
\begin{center}
\includegraphics[width=\columnwidth]{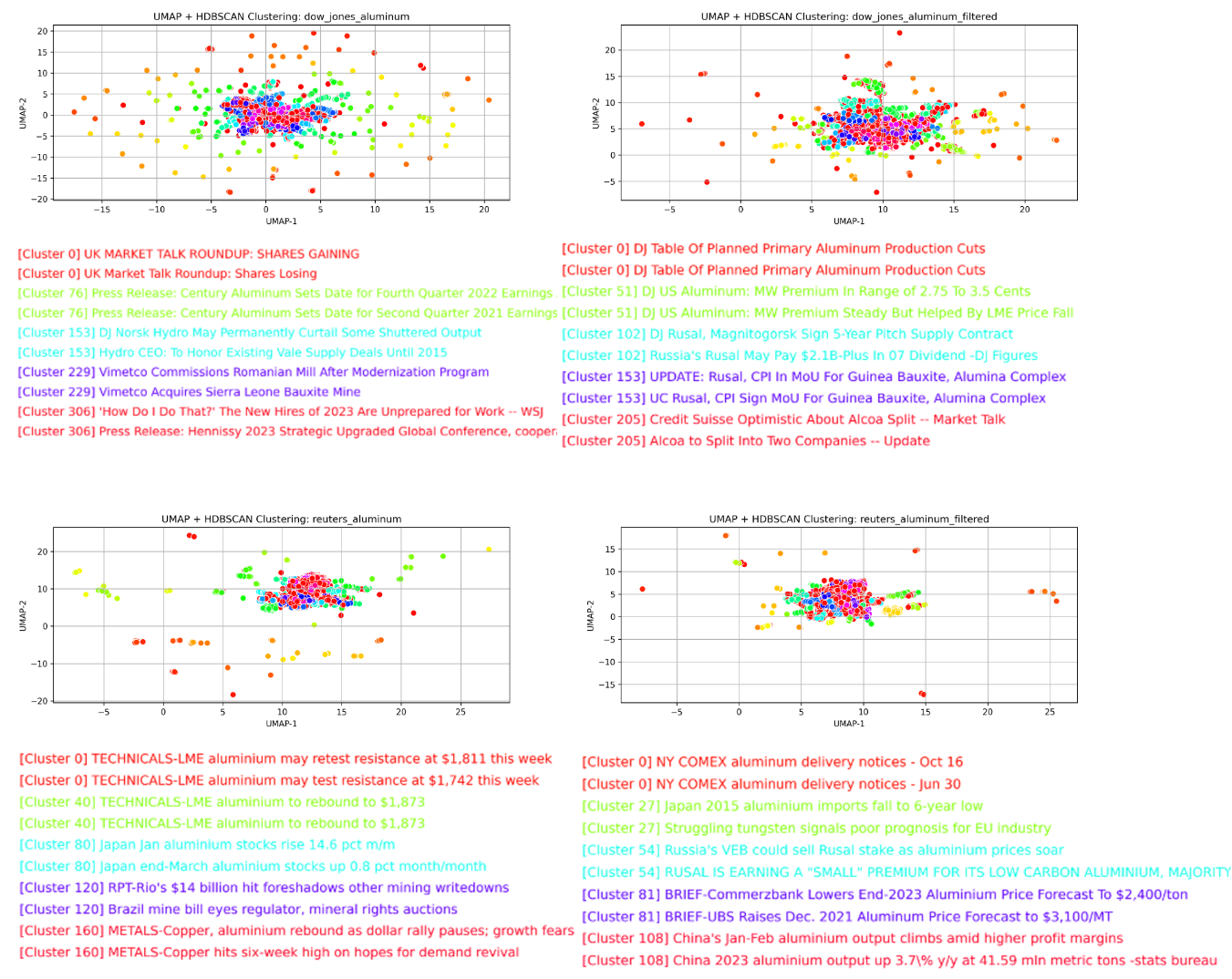}
\caption{Cluster plot of the Reuters and Dow Jones Newswires aluminum datasets before (left) and after (right) filtering using LLaMA3. The y and x axis are the 2 dimensions obtained by reducing the embeddings dimensions. The plots on the left show less number of outliers outside the main groups clusters.}
\label{fig:fig4}
\end{center}
\end{figure}

\section{Time Series Models Training and Testing} \label{sec:timeseries_training}
\begin{figure}[!ht]
\begin{center}
\includegraphics[width=\columnwidth]{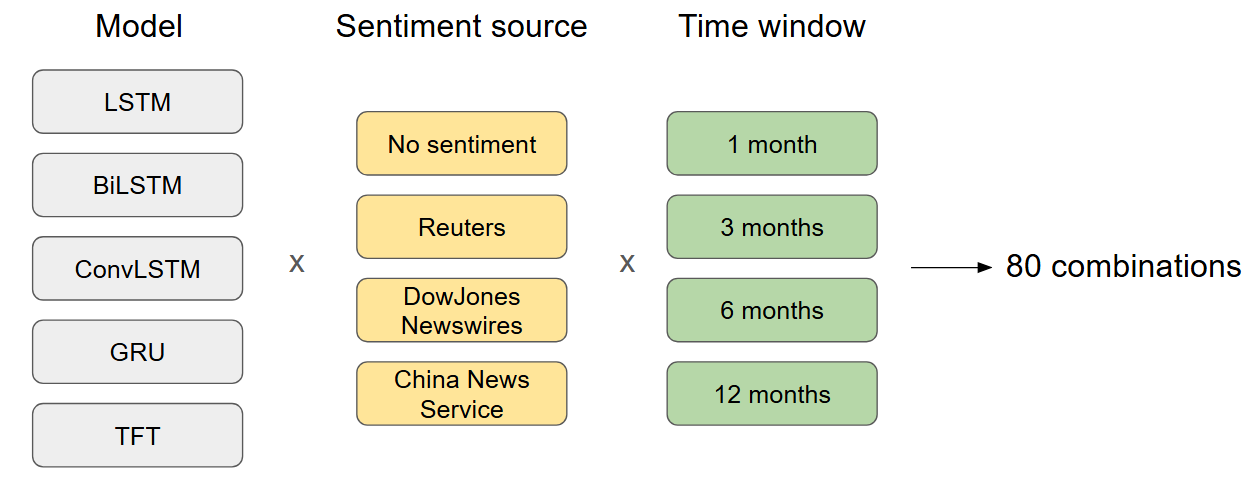}
\caption{Combination of models, sentiment sources and time windows}
\label{fig:fig12}
\end{center}
\end{figure}

The training methodology begins with data preprocessing, where the dataset containing metal prices and economic indicators is loaded, sorted by date, and missing values are imputed. The dataset is then resampled to generate monthly time series, adding sentiment score variables from each of the aluminum news sources (Reuters, DowJones Newswires and China News Service) and leaving one baseline dataset without sentiment variable. Rolling windows of 4 different lengths (1-month, 3-months, 6-months, and 12-months) are created to generate sequences of input features and corresponding target values, reshaped to match the input requirements of the models. This, added to the 5 different models and 4 sentiment sources, leaves us with 80 combinations, represented in Figure \ref{fig:fig12}. The models are trained using mean squared error loss and the Adam optimizer, with gradient clipping and learning rate scheduling applied to ensure stable convergence. 

The evaluation methodology uses walk-forward validation, a technique designed specifically for time series forecasting. In this approach, the model is trained on a rolling window of past observations, starting with an initial segment of the historical data. After training in this window, the model predicts the value immediately following the training period. The window then slides forward to include the new data point, and the process repeats iteratively for each subsequent time step. This method ensures that at each prediction, the model only has access to information that would have been available at that time, closely mimicking real-world forecasting scenarios and preventing data leakage, as well as maximizing the testing period. 

A grid search or hyperparameter search is conducted over hidden sizes, numbers of layers, and dropout values, with results recorded for each configuration. The chosen ranges are 16, 32, 64, 128 and 256 for hidden size and 1 to 6 for the number of layers. The dropout value used is always 0.1. Therefore, added to the original 80 setup combinations makes up a total of 2400 combinations of variables and parameters. However, only the best result for every hidden size and number of layers combination is collected, leaving 80 best results in total. The best-performing model for each window size is retrained on the full dataset and stored along with its predictions and evaluation metrics. This approach ensures robust model assessment, accounts for temporal dependencies, and minimizes data leakage in time series forecasting tasks.
 
The models performance are assessed using metrics such as R², RMSE and MAE. These are defined as
\begin{align}
\text{R}^2 &= 1 - \frac{\sum_{i=1}^{n} \left(y_i - \hat{y}_i\right)^2}{\sum_{i=1}^{n} \left(y_i - \bar{y}\right)^2} \\[2mm]
\text{RMSE} &= \sqrt{\frac{1}{n} \sum_{i=1}^{n} \left(y_i - \hat{y}_i\right)^2} \\[1mm]
\text{MAE} &= \frac{1}{n} \sum_{i=1}^{n} \left|y_i - \hat{y}_i\right|
\end{align}
where \(y_i\) denotes the true observed value for index \(i\), \(\hat{y}_i\) is the prediction of the corresponding model, \(\bar{y}\) represents the mean of the observed values, \(i\) indexes each observation and \(n\) is the total number of observations.

\section{Time Series Models Results} \label{sec:timeseries_training_results}
\subsection{Results for hidden size and number of layers from the grid search}

The best result for every combination of hidden size and number of layers is reported in Appendix \ref{grid_search_results_table} in Appendix B, reporting the best combination of hyperparameters.

\begin{figure}[!ht]
\begin{center}
\includegraphics[width=\columnwidth]{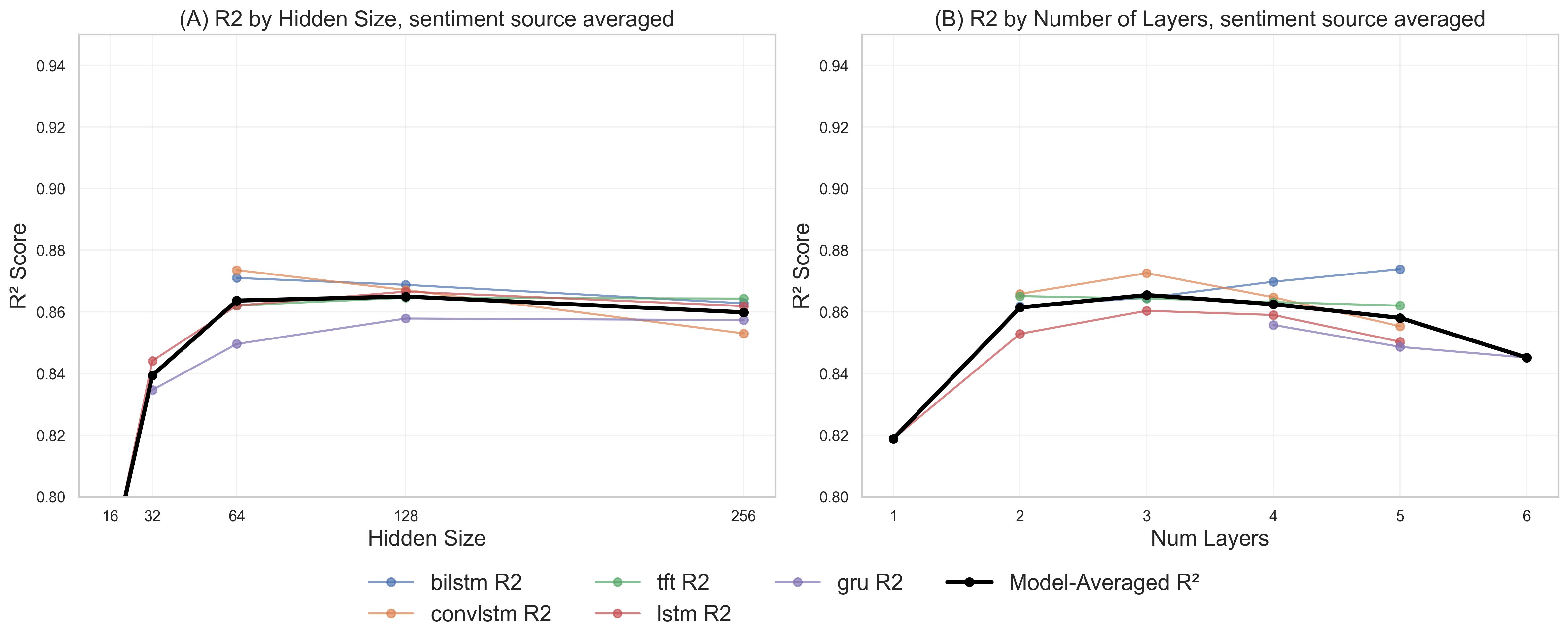}
\caption{Average $R^2$ score by number of layers and by hidden size. The solid line in black is the average of the five time series models. The faded lines are each respective model $R^2$, blue for BiLSTM, green for TFT, purple for GRU, orange for ConvLSTM and red for LSTM.}
\label{fig:fig14}
\end{center}
\end{figure}

Regarding hidden size and number of layers, we find that the average optimal values are around 3 layers and 128 hidden size, varying between models. These are shown in Figure \ref{fig:fig14}. A bigger number of layers tends to overfit, as well as hidden size values exceeding 128.

\subsection{Aluminum Price Forecasting Results}
\begin{figure}[!ht]
\begin{center}
\includegraphics[width=\columnwidth]{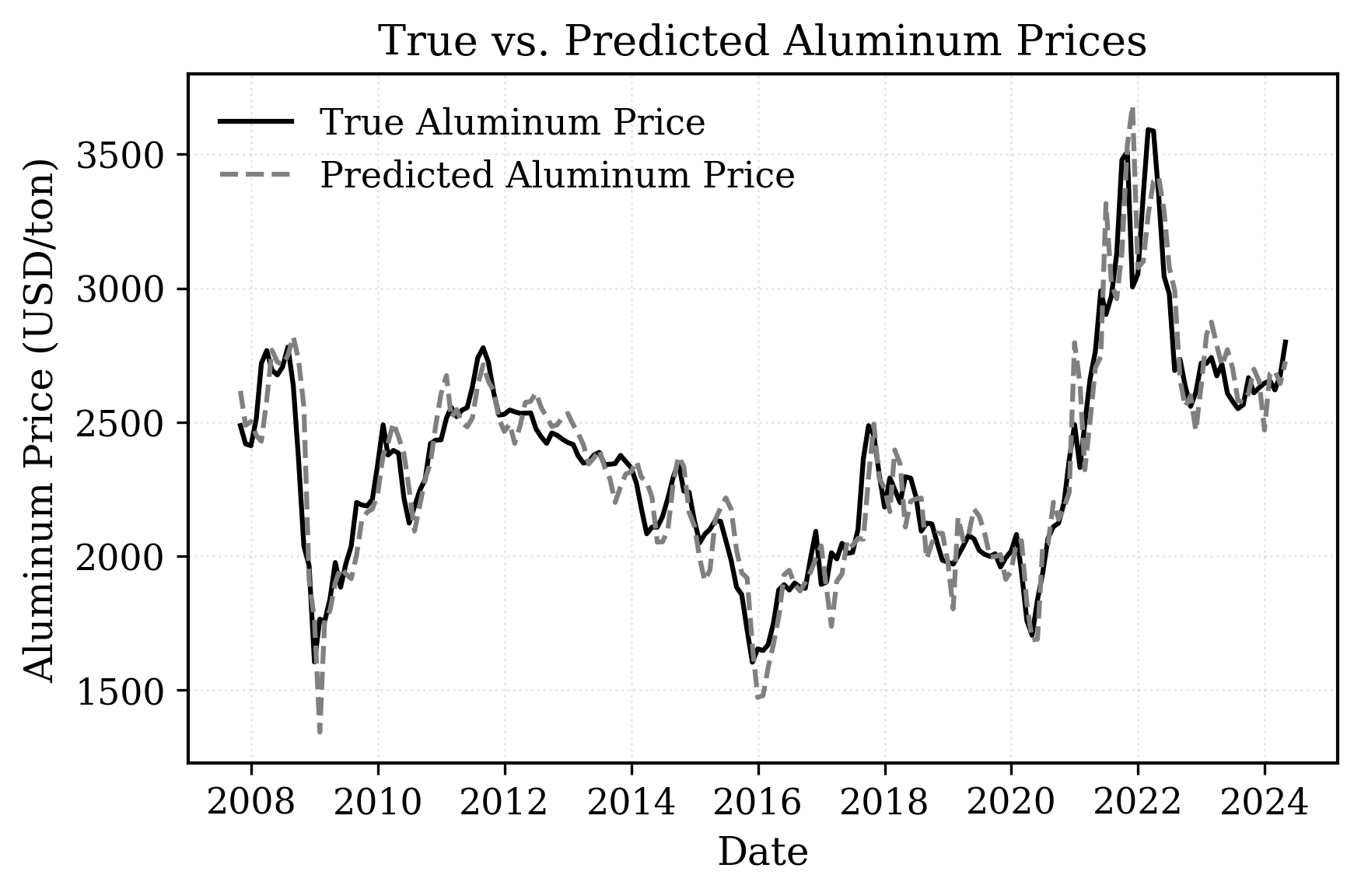}
\caption{Predicted aluminum price and true aluminum price from November 2007 to April 2024.}
\label{fig:fig13}
\end{center}
\end{figure}

Since the walk-forward validation method allows us to test with almost the whole dataset, we also gather the predictions from each best model, sentiment source, and time window. Figure \ref{fig:fig13} represents the observed aluminum price (solid line) and the aluminum price predicted by LSTM (dashed line) within a time window of 6-months and using sentiment scores from Reuters news.

\begin{figure*}[!ht]
\begin{center}
\includegraphics[width=\textwidth]{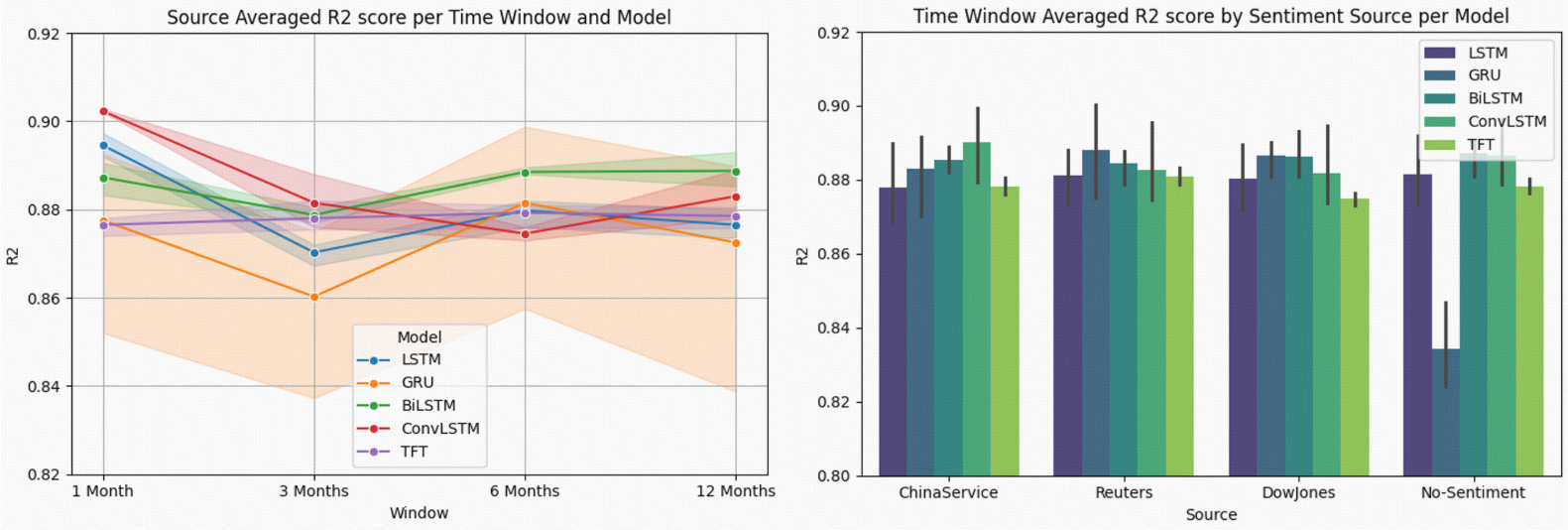}
\caption{Sentiment source averaged R$^2$ score per time window and model on the left, bands show the 95\% confidence intervals of those mean R² values. Time window averaged R$^2$ score per sentiment source and model on the right, whiskers show the 95\% confidence intervals of those mean R² values.}
\label{fig:fig15}
\end{center}
\end{figure*}

To study the impact of the sentiment variable and different time windows, we averaged the results by sentiment source and time window per model. These results are represented in Fig \ref{fig:fig15}.

The observed high predictive performance of models using short historical windows for next-month forecasting aligns with established principles in time series econometrics. Commodity prices exhibit time-varying autocorrelation structures where recent observations contain the strongest predictive signals, while distant historical data often introduce noise rather than information, a phenomenon documented in financial time series analysis \citep{hamilton1994timeseries}. This pattern is particularly pronounced in metal markets, where short-term dynamics is driven by recent supply-demand shocks and sentiment, while longer historical patterns reflect structural breaks and regime changes that provide limited incremental predictive power for immediate forecasts \citep{pindyck1999commodity}.

The preference for recent data over extended historical windows reflects optimal feature selection in non-stationary financial environments. As established in commodity forecasting research \citep{borovkova2011news_metals}, financial markets mix persistent fundamentals with transient noise, where short windows effectively isolate relevant signals while discarding historical noise that could dilute predictive accuracy. The observed improved performance of 6-month windows on both shorter (3-month) and longer (12-month) alternatives aligns with previous findings \citep{degiannakis2014multistep}, who demonstrate that intermediate historical windows optimally balance short-term noise filtering with sufficient context to capture business cycle patterns in metal markets. This intermediate-length window provides enough data to identify meaningful cyclical patterns while avoiding regime changes and structural breaks that increasingly contaminate longer historical series. This approach also mitigates overfitting to spurious patterns that appear significant in the sample but do not generalize, a critical concern in financial forecasting identified in empirical market studies \citep{lo2004adaptive}. The methodology aligns with evidence showing diminishing marginal information gains from additional historical data beyond intermediate windows \citep{inger2018optimal}, creating an effective trade-off between model complexity and forecasting robustness where 6-month windows can represent a sweet spot for aluminum price prediction.

The aggregated results reveal clear performance hierarchies and interaction effects between model architectures and sentiment sources. ConvLSTM emerges as the architecture that performs the best overall, achieving the highest average $R^2$ of 0.8902 when paired with ChinaService sentiment data, while also providing the strongest accuracy for monthly predictions (0.9022) when averaged across all sentiment sources. BiLSTM demonstrates remarkable consistency, maintaining nearly uniform performance across time windows (0.8788-0.8888) and excelling particularly with no-sentiment data (0.8870 average). GRU exhibits the greatest sensitivity to external information, showing a substantial 0.0538 performance gap between sentiment-enhanced configurations and the no-sentiment baseline, indicating strong dependency on external feature engineering. Interestingly, the no-sentiment baseline proves surprisingly competitive across multiple architectures, matching or even exceeding sentiment-augmented performance for LSTM, BiLSTM and ConvLSTM, suggesting that raw price history contains sufficient predictive signals for these models, while sentiment features can introduce conflicting noise rather than pure signal enhancement.

The analysis further uncovers distinct temporal performance patterns and source-model synergies. One-month prediction windows consistently yield the best results across architectures, supporting the temporal locality hypothesis, where recent data contains the strongest predictive signals. Reuters sentiment emerges as the most reliable information source, providing consistently strong performance across all models with minimal variability, while ChinaService shows specific synergy with ConvLSTM architecture. Performance declines with 3 months historical windows and grows back with 6 months windows.

\subsection{Portfolio results}
We apply the price-based trading strategy described in section \ref{section:trading_strategy} to the monthly aluminum price predictions made by the best hyperparameter combination found during the grid search for each type of model, sentiment source, and time window, gathered in Appendix \ref{grid_search_results_table}. 
Then we estimated R$^2$, RMSE, MAE, Hit Rate, p value, and total returns over the period of time in Appendix \ref{portfolio_results_table}. These metrics will be discussed in the following section.

\subsubsection{Hit rate}
Hit rate analysis provides a direct measure of a model's predictive quality, independent of position sizing, transaction costs, and market volatility effects. The hit rate refers to the percentage of forecasts in which the model correctly predicts the direction of price movement (up or down) over a given horizon. A consistently high hit rate (>0.50) demonstrates that the model captures meaningful directional signals beyond random chance, offering a systematic edge crucial for long-term trading viability. While profitability metrics like total returns and Sharpe ratios evaluate overall performance, hit rates reveal whether the strategy's success stems from genuine predictive power or merely from risk management and occasional large wins. This distinction is vital for strategy robustness, as models with low hit rates but positive returns often depend on unsustainable market conditions or excessive risk-taking, whereas high hit rates indicate reliable signal generation that can be scaled and optimized with proper execution.

The hit rate results reported in Table \ref{tab:hit_rate_performance} reveal a systematic hierarchy in predictive accuracy across model architectures and sentiment sources. The 3-month forecasting window consistently delivers peak performance across both models and sentiment sources, with Reuters sentiment achieving the highest overall hit rate of 0.577 at this horizon. Among architectures, ConvLSTM shows a better short-term forecasting capability with a 1-month hit rate of 0.585 and the highest model-average accuracy (0.564), while BiLSTM performs best at 3-months horizons. Notably, the no-sentiment baseline maintains competitive hit rates across all time windows and achieves a source-average of 0.551, only slightly below sentiment enhanced sources, suggesting that historical price patterns contain substantial directional information independent of sentiment signals. The observed stability in the hit rates on different horizons and configurations indicates robust predictive capabilities, with all configurations maintaining statistically significant accuracy above random chance and providing an overall average hit rate of 0.556 (55.6\%) regardless of window length or data source.

\begin{table}[!ht]
\centering
\resizebox{\columnwidth}{!}{%
\begin{tabular}{l l c c c c c}
\toprule
\multicolumn{2}{c}{\textbf{Configuration}} & \multicolumn{4}{c}{\textbf{Time Window}} & \\
\cmidrule(lr){3-6} \cmidrule{7-7}
\textbf{Category} & \textbf{Name} & \textbf{1M} & \textbf{3M} & \textbf{6M} & \textbf{12M} & \textbf{Avg} \\
\midrule

\multicolumn{7}{l}{\textbf{A. Model Performance}} \\
\midrule

\multirow{5}{*}{\rotatebox[origin=c]{90}{\textbf{Models}}}
& BiLSTM   & 0.549 & 0.573 & 0.562 & 0.534 & 0.554 \\
& ConvLSTM & 0.585 & 0.569 & 0.559 & 0.543 & \textbf{0.564} \\
& GRU      & 0.550 & 0.545 & 0.562 & 0.545 & 0.551 \\
& LSTM     & 0.564 & 0.563 & 0.549 & 0.548 & 0.556 \\
& TFT      & 0.554 & 0.568 & 0.538 & 0.560 & 0.555 \\
\cmidrule{2-7}
& \textbf{Model Avg} & 0.560 & \textbf{0.563} & 0.554 & 0.546 & 0.556 \\
\midrule

\multicolumn{7}{l}{\textbf{B. Sentiment Source Performance}} \\
\midrule

\multirow{4}{*}{\rotatebox[origin=c]{90}{\textbf{Sources}}}
& ChinaSvc & 0.570 & 0.567 & 0.558 & 0.550 & \textbf{0.561} \\
& DowJones & 0.551 & 0.549 & 0.555 & 0.554 & 0.552 \\
& No-Sent. & 0.564 & 0.561 & 0.543 & 0.536 & 0.551 \\
& Reuters  & 0.557 & 0.577 & 0.560 & 0.543 & 0.559 \\
\cmidrule{2-7}
& \textbf{Source Avg} & 0.560 & \textbf{0.563} & 0.554 & 0.546 & 0.556 \\
\bottomrule
\end{tabular}
}
\caption{Directional Hit Rate Performance by Model
and Sentiment Source Across Time Windows. Hit
rates represent the percentage of correct directional
predictions. Values closer to 1.000 indicate bet-
ter performance. The 3-month window shows the
highest average hit rate across both models and
sources. Values in bold highlight the time window
(panels A and B) with the best average hit rate, and
the model (panel A) and sentiment source (B) with
the best average hit rate.}
\label{tab:hit_rate_performance}
\end{table}

%

\subsubsection{Average returns}
The return performance analysis presented in Table \ref{tab:return_performance} reveals substantial heterogeneity in both the forecasting efficacy and the strategic value of the models evaluated. In particular, these results represent the average performance across all tested configurations rather than the maximum achievable performance under ideal settings. Consequently, many of the reported return metrics are negative or near zero, reflecting the dilution effect of including underperforming setups in the ensemble average. In Table \ref{tab:return_performance}A, the results shown are the sentiment averaged returns for each model and time window. LSTM performs best in the 1-month horizon, achieving a 168.7\% return (1.687 multiplicative factor), while GRU shows the best performance at 6-months time window with a 108.7\% return (1.087). GRU exhibits the most volatile performance profile, transitioning from -30.8\% at 3 months to 108.7\% at 6 months. TFT consistently underperforms on all time horizons, yielding predominantly negative returns and the lowest average performance of -42.9\% (-0.429). BiLSTM and ConvLSTM show mixed results, with BiLSTM achieving moderate success at shorter horizons (52.3\% at 3 months) but declining at longer horizons, while ConvLSTM shows minimal positive performance only at the 12-month horizon (14.7\%). The best performing models can reach high returns: LSTM (292\%), GRU (266\%) and BiLSTM (232\%). Despite their relatively high complexity, TFT and ConvLSTM tend to underperform, indicating that higher model complexity does not guarantee higher returns.

\begin{figure}[!ht]
\begin{center}
\includegraphics[width=\columnwidth]{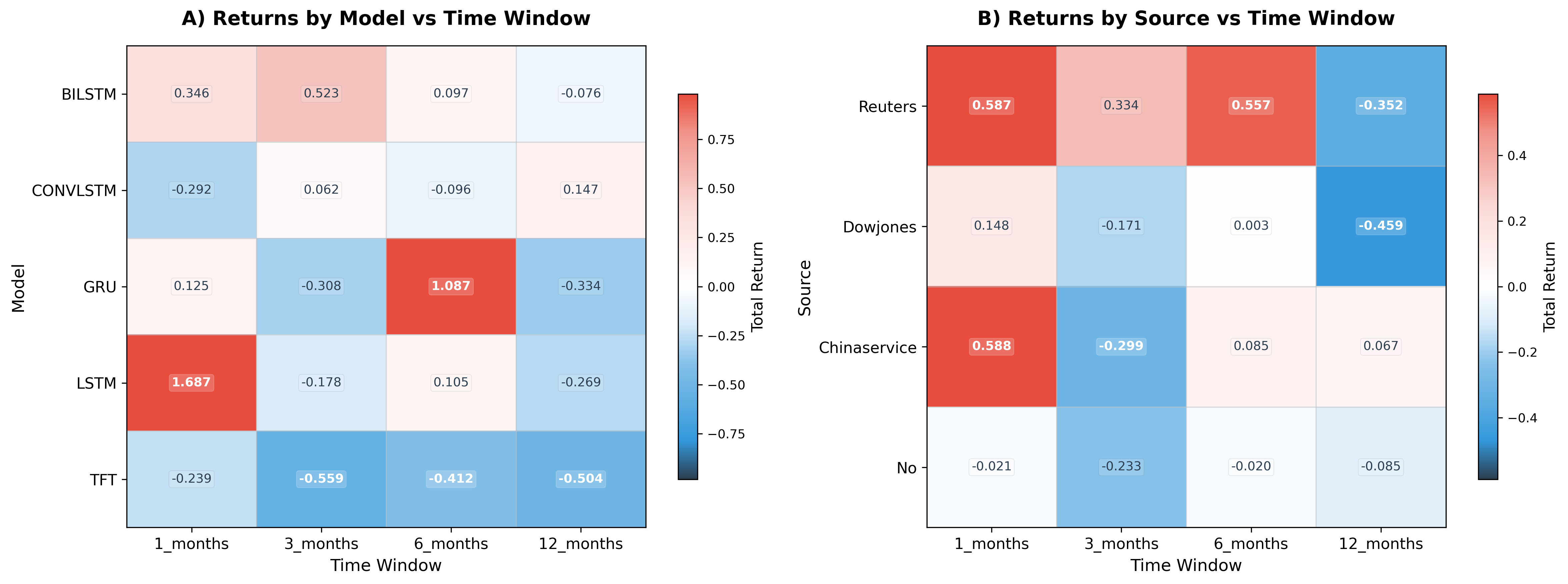}
\caption{Model averaged returns by time window (A). These are the total returns of all setups averaged for each model and time window. Source averaged returns (B) by time window. These are the total returns of all setups averaged for each sentiment source and time window.}
\label{fig:fig16}
\end{center}
\end{figure}

In Table \ref{tab:return_performance}B, the model averaged returns for each sentiment source and time window reveal that sentiment enhanced models generally outperform the no-sentiment baseline. Reuters sentiment emerges as the most effective source overall, achieving a 28.2\% average return and showing particularly strong medium-term performance with 55.7\% return at the 6-month horizon. The sentiment of ChinaService demonstrates strong short-term efficacy with 58.8\% return at 1 month, but experiences performance deterioration over longer horizons. In particular, all sentiment sources, including the baseline without sentiment, exhibit negative or almost negative performance at the 12-month horizon, with Dow Jones sentiment showing the most significant decline (-45.9\% return). This consistent pattern suggests that while sentiment integration can enhance short-to-medium term forecasting, its predictive value diminishes over longer horizons. The results, visualized in Figure \ref{fig:fig16}, show how the sentiment of Reuters peaks within a medium term window, while predictive accuracy remains systematically poor over longer time windows.

\begin{table}[!ht]
\begin{center}
\label{tab:return_performance}

\resizebox{\columnwidth}{!}{%
\begin{tabular}{l l c c c c c c}
\toprule
\multicolumn{2}{c}{\textbf{Configuration}} & \multicolumn{4}{c}{\textbf{Time Window}} & & \\
\cmidrule(lr){3-6} \cmidrule(lr){7-8}
\textbf{Category} & \textbf{Name} & \textbf{1M} & \textbf{3M} & \textbf{6M} & \textbf{12M} & \textbf{Avg} & \textbf{Best} \\
\midrule

\multicolumn{8}{l}{\textbf{A. Model Performance}} \\
\midrule

\multirow{5}{*}{\rotatebox[origin=c]{90}{\textbf{Models}}}
& BiLSTM   & 0.346 & 0.523 & 0.097 & -0.076 & 0.222 & 2.32 \\
& ConvLSTM & -0.292 & 0.062 & -0.096 & 0.147 & -0.045 & 1.02 \\
& GRU      & 0.125 & -0.308 & 1.087 & -0.334 & 0.142 & 2.66 \\
& LSTM     & \textbf{1.687} & -0.178 & 0.105 & -0.269 & \textbf{0.336} & \textbf{2.92} \\
& TFT      & -0.239 & -0.559 & -0.412 & -0.504 & -0.429 & 0.05 \\
\cmidrule{2-8}
& \textbf{Avg} & \textbf{0.325} & -0.092 & 0.156 & -0.207 & 0.045 & \\
\midrule

\multicolumn{8}{l}{\textbf{B. Sentiment Source Performance}} \\
\midrule

\multirow{4}{*}{\rotatebox[origin=c]{90}{\textbf{Sources}}}
& ChinaSvc   & \textbf{0.588} & -0.299 & 0.085 & 0.067 & 0.110 & 1.52 \\
& DowJones   & 0.148 & -0.171 & 0.003 & -0.459 & -0.120 & 1.36 \\
& No-Sent.   & -0.021 & -0.233 & -0.020 & -0.085 & -0.090 & 1.31 \\
& Reuters    & 0.587 & 0.334 & 0.557 & -0.352 & \textbf{0.282} & \textbf{2.92} \\
\cmidrule{2-8}
& \textbf{Avg} & \textbf{0.325} & -0.092 & 0.156 & -0.207 & 0.045 & \\
\bottomrule
\end{tabular}
}
\caption{Total return performance by model and sentiment source across time windows. \textbf{A.} Sentiment-source averaged returns for each mode. \textbf{B.} Model-averaged returns for each sentiment source. Values represent total returns (multiplicative factors, where a value of X indicates an X$\times$100\% return). The ``Avg'' column shows time-window averaged performance, and ``Best'' reports the highest return achieved.}
\end{center}
\end{table}

The relationship between directional accuracy (hit rate) and realized returns reveals a complex association between predictive quality, given by the hit rate, and financial performance, given by the returns. ConvLSTM, although it delivers the best average hit rate, achieves the second lowest average returns (-4.5\%). This divergence suggests that while hit rates provide fundamental predictive validation, and therefore all models maintain statistically significant accuracy above random, financial performance requires additional mechanisms beyond directional correctness. The 1-month forecast window consistently optimizes both metrics, achieving the second best average hit rate (0.560) and highest average return (32.5\%), indicating that the latest signals are most important when predicting future prices. Sentiment addition similarly enhances both dimensions, with all sentiment sources delivering a higher average hit rate, higher returns, and higher best returns. In particular, the best performing setups are the ones that add sentiment scores from Reuters.

\subsubsection{Best portfolio performance from no-sentiment vs sentiment leveraged model}
While the previous results show how adding sentiment from either Reuters, Dow Jones or China News Service improves the model predictions $R^2$, hit rate and total returns across the whole time interval of study (2007 - 2024) when averaged by sentiment source or model, this section compares the results of the configuration of hyperparameters, model type and time window, which gives the best total returns without using sentiment scores and when sentiment scores from a source (Reuters, Dow Jones or China News Service) are used. 

Without the use of sentiment scores, which we refer to as ``no-sentiment'' configuration, the best total returns were achieved with the predictions made by the LSTM model in a 1-month time window, 128 hidden size and 4 layers. This model achieved a $R^2$ score of 0.90, RMSE of 120.17, and MAE of 84.28. After applying the same simulation of the trading strategy as described in \ref{section:trading_strategy} to its predicted monthly aluminum prices, it achieved a total return of 131\%. 

However, when including sentiment, the highest total returns were achieved by the configuration of the LSTM model in a 1-month time window, 128 hidden size, and 4 layers with sentiment from Reuters. Initially, it got an $R^2$ score of 0.89, RMSE of 125.12, and MAE of 87.98, which is slightly worse than the no-sentiment variant, whereas after running the trading simulation the total returns outperform, achieving a total return of 292\%.

The substantial increase in total returns (292\% vs 131\%) observed when incorporating Reuters sentiment, despite marginally lower $R^2$, RMSE and MAE scores, can be attributed to the directional accuracy and timing precision of the model’s forecasts, which are not fully captured by point-prediction error metrics.

$R^2$, RMSE and MAE measure the magnitude of prediction errors in all time steps, penalizing deviations in predicted price levels regardless of whether the forecast correctly anticipates the direction of price movement. In contrast, trading performance depends critically on correctly predicting price direction at key turning points—particularly around market reversals, news events, or sentiment shifts.

Although the no-sentiment model can achieve slightly better average error metrics, the Reuters-augmented model appears to generate more accurate signals at economically meaningful moments, such as identifying impending price increases or decreases that lead to profitable trading decisions. This suggests that sentiment information helps the model better align its predictions with market-moving events reflected in the news flow, even if it introduces small increases in the average prediction error.

\begin{figure}[!ht]
\begin{center}
\includegraphics[width=\columnwidth]{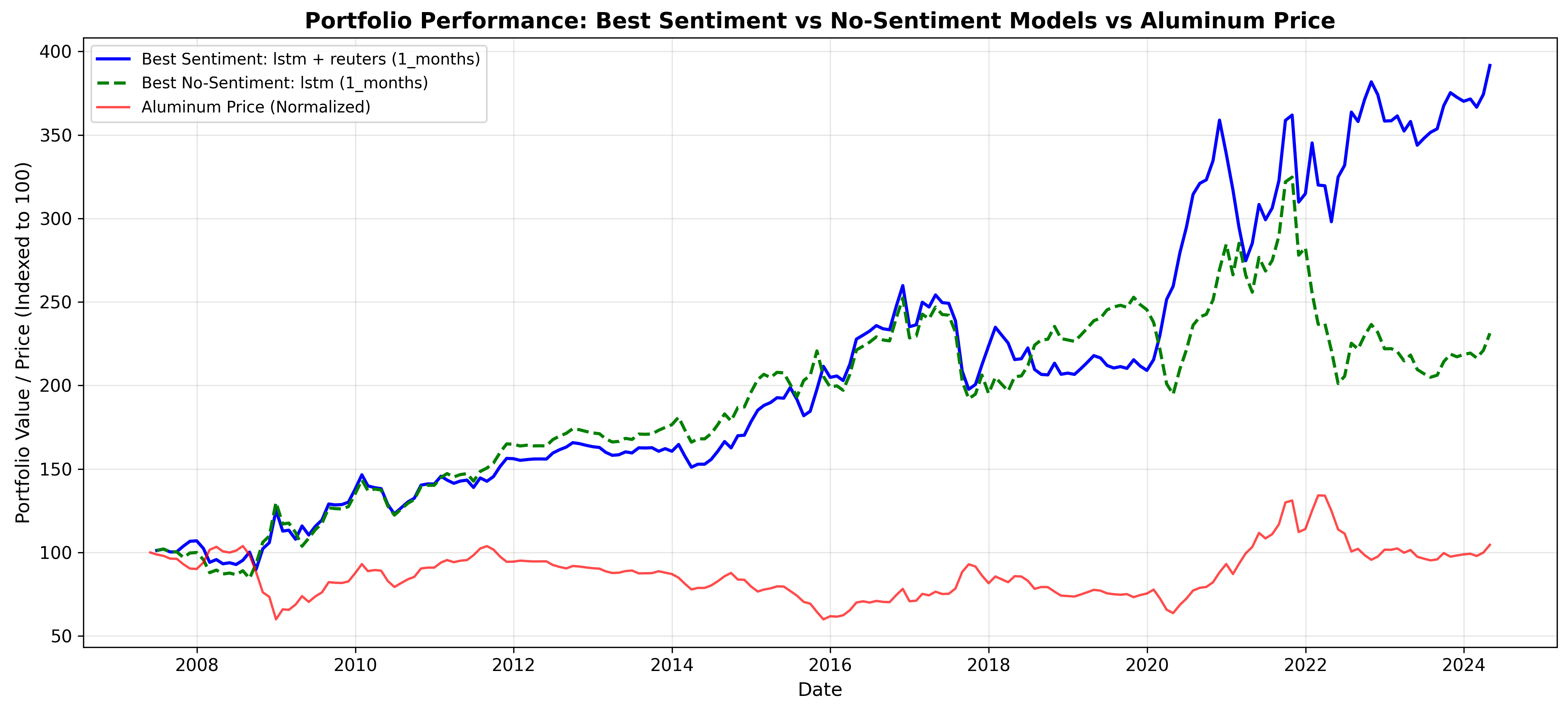}
\caption{Comparison of the best Sentiment vs No-Sentiment Models returns vs True Aluminum price. Y axis is the Portfolio Value, with the starting point being 100 (100\% of the intial value).}
\label{fig:fig17}
\end{center}
\end{figure}

\begin{figure}[!ht]
\begin{center}
\includegraphics[width=\columnwidth]{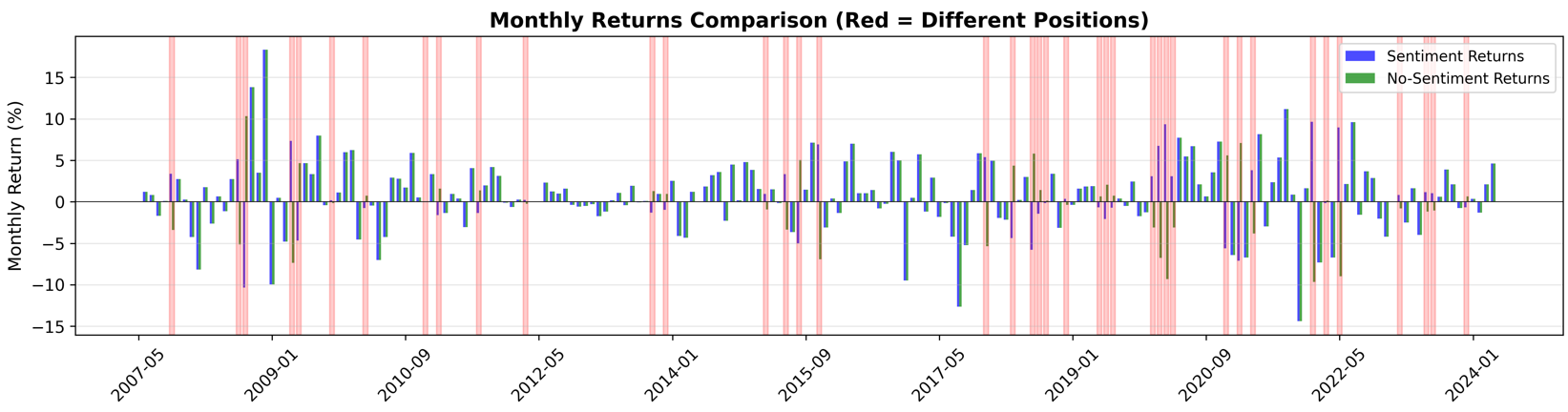}
\caption{Monthly return of best sentiment and no-sentiment models monthly returns. The red bars show month where the two configurations made a different decision.}
\label{fig:fig18}
\end{center}
\end{figure}

Figure \ref{fig:fig17} represents both the portfolio performance over time and the evolution of the aluminum price. Although both follow a similar pattern, the configuration that adds Reuters sentiment outperforms due to several different trade decisions. Figure \ref{fig:fig18} shows the monthly return comparison between both variants. Most of the time, the signals are the same for both configurations (with and without sentiment); however, whenever the configurations disagree, the model with sentiment is more often correct. The red lines represent a signal disagreement, where the sentiment and no-sentiment models execute a different trade. Some of the most determinant trades were executed between the 2020 and 2022 time periods. The three largest returns gaps between both models were in March 2020, January 2022, and can 2022.

To illustrate the results, take month February 2020, with an observed aluminum price of 1940.50\$/ton. While the no-sentiment model predicted a 1974.37\$/ton price for March and the reuters sentiment model predicted a 1899.57\$/ton, with an observed price for March 2020 of 1759.43\$/ton. Here the long signal captured by the model without sentiment leads to a -9.33\% loss in returns, whereas the short signal captured by the sentiment model generated a 9.33\% return.

Looking at the sentiment variable that was used to make the predictions, the Reuters sentiment score for February 2020 took the lowest possible value (-1), which is expected to decrease the predicted price value for next month. Table \ref{tab:sentiment_compact} shows the headline, date and associated sentiment generated by the finetuned language model (SFT Qwen3 8B) for February 2020.
\definecolor{negred}{RGB}{255,230,230}
\definecolor{posgreen}{RGB}{230,255,230}
\definecolor{neutyellow}{RGB}{255,255,230}

\begin{table}[!ht]
\begin{center}
\label{tab:sentiment_compact}

\resizebox{\columnwidth}{!}{%
\begin{tabular}{p{0.65\textwidth} c c}
\toprule
\textbf{News Headline} & \textbf{Date} & \textbf{Sentiment} \\
\midrule
Coronavirus will negatively affect aluminum market in China & 28-Feb-20 & \cellcolor{negred}Negative \\
Coronavirus double shock for aluminium sector & 24-Feb-20 & \cellcolor{negred}Negative \\
LME aluminium can test support at \$1,688 & 24-Feb-20 & \cellcolor{negred}Negative \\
Japan aluminium stocks down 2.8\% & 18-Feb-20 & \cellcolor{negred}Negative \\
LME aluminium testing support at \$1,709 & 10-Feb-20 & \cellcolor{negred}Negative \\
LME aluminium seeking support at \$1,709 & 03-Feb-20 & \cellcolor{negred}Negative \\
Shanghai metals limit-down amid coronavirus fears & 03-Feb-20 & \cellcolor{negred}Negative \\
\bottomrule
\end{tabular}
}
\caption{Sentiment analysis of market news (February 2020).}
\end{center}
\end{table}

Let us analyze January 2022 as a second illustration of the results. The model without sentiment model had predicted 3044.21\$/ton while sentiment model predicted 3088.49\$/ton. The initial price on December 2021 was 3052.88\$/ton and the price on January 2022 surged to 3347.41\$/ton. Therefore, the no-sentiment model suggested a short trade while the sentiment one suggested a long trade, leading to -9.64\% and +9.64\% return respectively. The predictions made by the finetuned Qwen3 model, on December of 2021, predicted a sentiment score of 0.18, slightly positive. Table \ref{tab:sentiment_dec2021} shows all headlines, data, and associated sentiment for the investigated month.
\definecolor{negred}{RGB}{255,230,230}
\definecolor{posgreen}{RGB}{230,255,230}
\definecolor{neutyellow}{RGB}{255,255,230}

\begin{table}[!ht]
\begin{center}
\label{tab:sentiment_dec2021}

\resizebox{\columnwidth}{!}{%
\begin{tabular}{p{0.65\textwidth} c c}
\toprule
\textbf{News Headline} & \textbf{Date} & \textbf{Sentiment} \\
\midrule
Norway's Hydro to cut Slovakia aluminium output further due to power prices & 30-Dec-21 & \cellcolor{negred}Negative \\
Copper slips in range-bound trade, aluminium shines on supply worries & 30-Dec-21 & \cellcolor{negred}Negative \\
Power price surge pushes aluminium to 2-month high & 23-Dec-21 & \cellcolor{posgreen}Positive \\
China Nov aluminium output at 3.10 mln tonnes -- stats bureau & 15-Dec-21 & \cellcolor{neutyellow}Neutral \\
Scarce supplies to propel aluminium to top LME leaderboard & 15-Dec-21 & \cellcolor{posgreen}Positive \\
Marubeni sees Japanese aluminium premiums at \$140--\$250/T in 2022 & 07-Dec-21 & \cellcolor{neutyellow}Neutral \\
Aluminium prices firm as China plans hiking coal contract prices & 03-Dec-21 & \cellcolor{posgreen}Positive \\
Japan aluminium stocks in October up 1.1\% m/m -- Marubeni & 03-Dec-21 & \cellcolor{posgreen}Positive \\
London aluminium edges higher as stockpiles fall, demand recovers & 02-Dec-21 & \cellcolor{posgreen}Positive \\
Aluminium dips on Omicron fears, but low inventories cushion fall & 02-Dec-21 & \cellcolor{negred}Negative \\
Carbon brakes aluminium supply response to booming prices: Andy Home & 01-Dec-21 & \cellcolor{neutyellow}Neutral \\
\bottomrule
\end{tabular}
}
\caption{Sentiment analysis of market news (December 2021).}
\end{center}
\end{table}

As a last example, let us consider April 2022, with an initial price of 3345.02\$/ton. While the no-sentiment model predicted a 3374.74\$/ton price for can the Reuters-based sentiment model predicted a 3326.42\$/ton price. In the end, the real price for March was 3044.82\$/ton. This triggered a long signal for the no-sentiment variant, losing -8.97\% whereas the sentiment model captured a short signal, earning a positive amount of 8.97\%.

On April 20, 2022, the sentiment score was -0.625, which is a relatively strong negative value. Table \ref{tab:sentiment_apr2022} shows the news headline, date and associated value for that month.

\begin{table}[!ht]
\begin{center}
\label{tab:sentiment_apr2022}

\resizebox{\columnwidth}{!}{%
\begin{tabular}{p{0.65\textwidth} c c}
\toprule
\textbf{News Headline} & \textbf{Date} & \textbf{Sentiment} \\
\midrule
London aluminium poised for worst month in over a decade on growth risks & 29-Apr-22 & \cellcolor{negred}Negative \\
China Shenhuo to raise aluminium output in Yunnan as power curbs ease & 26-Apr-22 & \cellcolor{posgreen}Positive \\
Global aluminium output falls 1.55\% in March year on year, IAI says & 20-Apr-22 & \cellcolor{negred}Negative \\
Shanghai aluminium hits 3-month low on strong dollar, demand woes & 12-Apr-22 & \cellcolor{negred}Negative \\
Shanghai aluminium sinks to 3-month low as demand woes linger & 12-Apr-22 & \cellcolor{negred}Negative \\
China demand angst hits aluminium prices & 11-Apr-22 & \cellcolor{negred}Negative \\
Shanghai aluminium slips to over 3-week low as demand concerns weigh & 08-Apr-22 & \cellcolor{negred}Negative \\
Japan aluminium buyers to pay lower premiums for April--June imports & 07-Apr-22 & \cellcolor{negred}Negative \\
Japan buyers agree to Q2 aluminium premium of \$172/T, sources say & 07-Apr-22 & \cellcolor{neutyellow}Neutral \\
\bottomrule
\end{tabular}
}
\caption{Sentiment analysis of market news (April 2022).}
\end{center}
\end{table}

\onecolumn

\section{Defined Volatility Regimes Across Time} \label{sec:volatility_regimes_figure}
\begin{figure*}[!ht]
\begin{center}
\includegraphics[width=\textwidth]{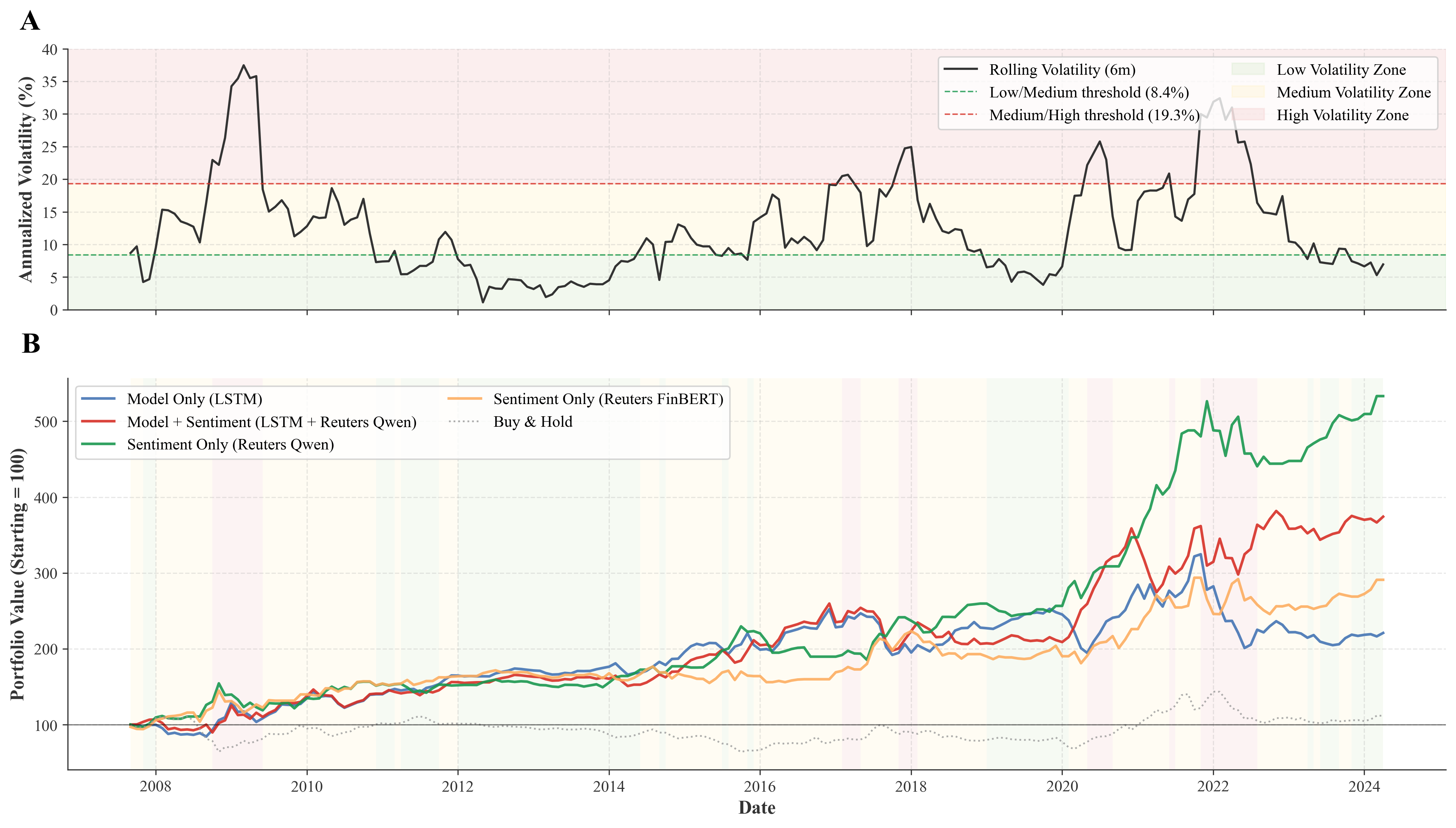}
\caption{Panel A: Volatility Regimes Over Time. Red area represents the high-volatility regime, yellow represents medium-volatility and green low-volatility. Panel B: Portfolio Values Over Time for the different strategies. Blue line represents the price-based strategy, red line price-based strategy combined with sentiment, green line sentiment only strategy with sentiment predicted by finetuned Qwen, yellow line sentiment only strategy with sentiment predicted by FinBERT and black line buy \& hold strategy.}
\label{fig:fig32}
\end{center}
\end{figure*}

\section{Grid Search best results per model, time window and sentiment source} \label{grid_search_results_table}

{\scriptsize  
\setlength{\tabcolsep}{2pt}
\renewcommand{\arraystretch}{0.9}

\begin{longtable}{l l l c c c c c}
\toprule
Model & Source & Horizon & Hidden Size & Layers & R$^2$ & RMSE & MAE \\
\midrule
\endfirsthead

\toprule
Model & Source & Horizon & Hidden Size & Layers & R$^2$ & RMSE & MAE \\
\midrule
\endhead

\midrule
\multicolumn{8}{r}{Continues on next page} \\
\midrule
\endfoot

\bottomrule
\endlastfoot

bilstm & reuters & 1 months & 256 & 5 & 0.89 & 126.79 & 91.59 \\
bilstm & reuters & 3 months & 256 & 3 & 0.88 & 132.66 & 92.02 \\
bilstm & reuters & 6 months & 64 & 3 & 0.89 & 126.92 & 88.74 \\
bilstm & reuters & 12 months & 64 & 3 & 0.89 & 128.64 & 95.47 \\
bilstm & dowjones & 1 months & 256 & 5 & 0.88 & 130.12 & 93.37 \\
bilstm & dowjones & 3 months & 128 & 4 & 0.88 & 129.93 & 92.71 \\
bilstm & dowjones & 6 months & 128 & 2 & 0.89 & 126.94 & 88.22 \\
bilstm & dowjones & 12 months & 128 & 5 & 0.89 & 123.87 & 91.57 \\
bilstm & chinaservice & 1 months & 64 & 5 & 0.89 & 125.11 & 89.85 \\
bilstm & chinaservice & 3 months & 256 & 5 & 0.88 & 130.69 & 95.01 \\
bilstm & chinaservice & 6 months & 256 & 5 & 0.89 & 126.80 & 96.00 \\
bilstm & chinaservice & 12 months & 128 & 2 & 0.88 & 130.00 & 98.25 \\
bilstm & no-sentiment & 1 months & 64 & 4 & 0.89 & 124.68 & 90.66 \\
bilstm & no-sentiment & 3 months & 64 & 3 & 0.88 & 131.74 & 89.92 \\
bilstm & no-sentiment & 6 months & 128 & 4 & 0.89 & 125.87 & 89.84 \\
bilstm & no-sentiment & 12 months & 128 & 4 & 0.89 & 127.39 & 91.64 \\
convlstm & reuters & 1 months & 64 & 3 & 0.90 & 117.84 & 82.64 \\
convlstm & reuters & 3 months & 128 & 3 & 0.88 & 132.64 & 90.43 \\
convlstm & reuters & 6 months & 64 & 3 & 0.87 & 134.80 & 96.54 \\
convlstm & reuters & 12 months & 64 & 3 & 0.88 & 132.48 & 97.26 \\
convlstm & dowjones & 1 months & 64 & 3 & 0.90 & 117.84 & 82.64 \\
convlstm & dowjones & 3 months & 128 & 3 & 0.88 & 132.64 & 90.43 \\
convlstm & dowjones & 6 months & 64 & 3 & 0.87 & 134.80 & 96.54 \\
convlstm & dowjones & 12 months & 64 & 3 & 0.88 & 134.74 & 97.52 \\
convlstm & chinaservice & 1 months & 64 & 3 & 0.90 & 117.84 & 82.64 \\
convlstm & chinaservice & 3 months & 64 & 3 & 0.89 & 124.18 & 87.29 \\
convlstm & chinaservice & 6 months & 128 & 5 & 0.87 & 134.19 & 97.63 \\
convlstm & chinaservice & 12 months & 64 & 3 & 0.89 & 125.63 & 93.95 \\
convlstm & no-sentiment & 1 months & 64 & 3 & 0.90 & 117.59 & 82.30 \\
convlstm & no-sentiment & 3 months & 128 & 3 & 0.88 & 129.37 & 89.36 \\
convlstm & no-sentiment & 6 months & 64 & 4 & 0.88 & 132.81 & 96.95 \\
convlstm & no-sentiment & 12 months & 128 & 5 & 0.88 & 130.05 & 96.46 \\
gru & reuters & 1 months & 64 & 6 & 0.89 & 126.34 & 89.13 \\
gru & reuters & 3 months & 64 & 4 & 0.87 & 135.72 & 98.21 \\
gru & reuters & 6 months & 256 & 4 & 0.90 & 117.66 & 86.79 \\
gru & reuters & 12 months & 128 & 6 & 0.89 & 127.26 & 97.05 \\
gru & dowjones & 1 months & 64 & 6 & 0.89 & 125.94 & 89.13 \\
gru & dowjones & 3 months & 128 & 4 & 0.88 & 131.83 & 92.54 \\
gru & dowjones & 6 months & 256 & 4 & 0.89 & 125.57 & 89.29 \\
gru & dowjones & 12 months & 128 & 4 & 0.89 & 126.84 & 91.94 \\
gru & chinaservice & 1 months & 128 & 6 & 0.89 & 122.44 & 90.03 \\
gru & chinaservice & 3 months & 128 & 4 & 0.87 & 138.04 & 98.17 \\
gru & chinaservice & 6 months & 128 & 4 & 0.88 & 129.52 & 91.20 \\
gru & chinaservice & 12 months & 256 & 4 & 0.89 & 127.25 & 92.13 \\
gru & no-sentiment & 1 months & 256 & 6 & 0.84 & 150.85 & 109.49 \\
gru & no-sentiment & 3 months & 256 & 4 & 0.83 & 157.38 & 111.45 \\
gru & no-sentiment & 6 months & 256 & 4 & 0.85 & 147.14 & 112.09 \\
gru & no-sentiment & 12 months & 128 & 4 & 0.82 & 161.15 & 125.29 \\
lstm & reuters & 1 months & 128 & 4 & 0.89 & 125.12 & 87.98 \\
lstm & reuters & 3 months & 128 & 2 & 0.87 & 135.41 & 94.14 \\
lstm & reuters & 6 months & 256 & 3 & 0.88 & 129.96 & 96.39 \\
lstm & reuters & 12 months & 256 & 2 & 0.88 & 131.37 & 92.85 \\
lstm & dowjones & 1 months & 128 & 3 & 0.90 & 122.05 & 86.22 \\
lstm & dowjones & 3 months & 256 & 3 & 0.87 & 135.47 & 98.11 \\
lstm & dowjones & 6 months & 64 & 2 & 0.88 & 129.94 & 95.99 \\
lstm & dowjones & 12 months & 256 & 2 & 0.87 & 135.88 & 97.83 \\
lstm & chinaservice & 1 months & 128 & 5 & 0.90 & 121.83 & 87.18 \\
lstm & chinaservice & 3 months & 64 & 3 & 0.87 & 138.20 & 100.94 \\
lstm & chinaservice & 6 months & 256 & 3 & 0.87 & 134.75 & 99.10 \\
lstm & chinaservice & 12 months & 64 & 2 & 0.88 & 134.19 & 98.15 \\
lstm & no-sentiment & 1 months & 128 & 4 & 0.90 & 120.17 & 84.28 \\
lstm & no-sentiment & 3 months & 64 & 2 & 0.87 & 134.49 & 96.65 \\
lstm & no-sentiment & 6 months & 64 & 2 & 0.88 & 130.91 & 95.83 \\
lstm & no-sentiment & 12 months & 256 & 2 & 0.87 & 135.68 & 97.44 \\
tft & reuters & 1 months & 128 & 5 & 0.88 & 131.63 & 95.42 \\
tft & reuters & 3 months & 128 & 3 & 0.88 & 128.74 & 90.06 \\
tft & reuters & 6 months & 128 & 4 & 0.88 & 130.66 & 92.73 \\
tft & reuters & 12 months & 256 & 3 & 0.88 & 131.93 & 93.59 \\
tft & dowjones & 1 months & 256 & 4 & 0.87 & 134.40 & 95.62 \\
tft & dowjones & 3 months & 128 & 5 & 0.88 & 132.15 & 96.22 \\
tft & dowjones & 6 months & 256 & 3 & 0.87 & 134.51 & 98.03 \\
tft & dowjones & 12 months & 128 & 4 & 0.88 & 134.85 & 96.60 \\
tft & chinaservice & 1 months & 64 & 3 & 0.88 & 132.26 & 98.11 \\
tft & chinaservice & 3 months & 256 & 5 & 0.87 & 133.48 & 96.20 \\
tft & chinaservice & 6 months & 128 & 2 & 0.88 & 130.66 & 96.20 \\
tft & chinaservice & 12 months & 64 & 2 & 0.88 & 132.34 & 96.14 \\
tft & no-sentiment & 1 months & 128 & 2 & 0.88 & 131.48 & 94.61 \\
tft & no-sentiment & 3 months & 64 & 2 & 0.88 & 132.73 & 93.59 \\
tft & no-sentiment & 6 months & 64 & 2 & 0.88 & 130.88 & 96.12 \\
tft & no-sentiment & 12 months & 256 & 4 & 0.88 & 133.36 & 96.45 \\
\bottomrule
\addlinespace[2ex]
\caption{Best results ($R^2$, RMSE and MAE) for the best hyperparameter combination for each model, sentiment source and time window.}
\label{tab:full_results}
\end{longtable}

}

\section{Portfolio results for every model, sentiment source and time window combination best grid search combination} \label{portfolio_results_table}

{\scriptsize
\setlength{\tabcolsep}{2pt}
\renewcommand{\arraystretch}{0.9}
\begin{longtable}{l l l r r r r r r}
\toprule
Model & Source & Horizon & R\textsuperscript{2} & RMSE & MAE & HitRate & p\_value & TotRet \\
\midrule
\endfirsthead
\toprule
Model & Source & Horizon & R\textsuperscript{2} & RMSE & MAE & HitRate & p\_value & TotRet \\
\midrule
\endhead
\midrule
\multicolumn{9}{r}{Continued on next page} \\
\midrule
\endfoot
\bottomrule
\endlastfoot
lstm & chinaservice & 1\_months & 0.90 & 121.83 & 87.18 & 0.56 & 0.0665 & 1.16 \\
lstm & chinaservice & 3\_months & 0.87 & 138.20 & 100.94 & 0.56 & 0.0651 & -0.60 \\
lstm & chinaservice & 6\_months & 0.87 & 134.75 & 99.10 & 0.55 & 0.1760 & -0.08 \\
lstm & chinaservice & 12\_months & 0.88 & 134.19 & 98.15 & 0.55 & 0.1283 & -0.21 \\
lstm & reuters & 1\_months & 0.89 & 125.12 & 87.98 & 0.55 & 0.1594 & 2.92 \\
lstm & reuters & 3\_months & 0.87 & 135.41 & 94.14 & 0.60 & 0.0041 & 0.45 \\
lstm & reuters & 6\_months & 0.88 & 129.96 & 96.39 & 0.53 & 0.3557 & 1.09 \\
lstm & reuters & 12\_months & 0.88 & 131.37 & 92.85 & 0.55 & 0.1283 & 0.17 \\
lstm & dowjones & 1\_months & 0.90 & 122.05 & 86.22 & 0.56 & 0.0906 & 1.36 \\
lstm & dowjones & 3\_months & 0.87 & 135.47 & 98.11 & 0.53 & 0.3977 & -0.39 \\
lstm & dowjones & 6\_months & 0.88 & 129.94 & 95.99 & 0.53 & 0.3557 & 0.05 \\
lstm & dowjones & 12\_months & 0.87 & 135.88 & 97.83 & 0.55 & 0.1693 & -0.62 \\
lstm & no-sentiment & 1\_months & 0.90 & 120.17 & 84.28 & 0.58 & 0.0158 & 1.31 \\
lstm & no-sentiment & 3\_months & 0.87 & 134.49 & 96.65 & 0.56 & 0.0890 & -0.18 \\
lstm & no-sentiment & 6\_months & 0.88 & 130.91 & 95.83 & 0.58 & 0.0177 & -0.64 \\
lstm & no-sentiment & 12\_months & 0.87 & 135.68 & 97.44 & 0.53 & 0.3483 & -0.42 \\
gru & chinaservice & 1\_months & 0.89 & 122.44 & 90.03 & 0.56 & 0.0906 & 1.52 \\
gru & chinaservice & 3\_months & 0.87 & 138.04 & 98.17 & 0.54 & 0.2035 & -0.48 \\
gru & chinaservice & 6\_months & 0.88 & 129.52 & 91.20 & 0.57 & 0.0534 & 0.36 \\
gru & chinaservice & 12\_months & 0.89 & 127.25 & 92.13 & 0.54 & 0.2193 & 0.05 \\
gru & reuters & 1\_months & 0.89 & 126.34 & 89.13 & 0.54 & 0.2611 & -0.19 \\
gru & reuters & 3\_months & 0.87 & 135.72 & 98.21 & 0.54 & 0.2035 & -0.57 \\
gru & reuters & 6\_months & 0.90 & 117.66 & 86.79 & 0.61 & 0.0018 & 2.66 \\
gru & reuters & 12\_months & 0.89 & 127.26 & 97.05 & 0.51 & 0.7188 & -0.71 \\
gru & dowjones & 1\_months & 0.89 & 125.94 & 89.13 & 0.55 & 0.1594 & -0.28 \\
gru & dowjones & 3\_months & 0.88 & 131.83 & 92.54 & 0.55 & 0.1573 & 0.26 \\
gru & dowjones & 6\_months & 0.89 & 125.57 & 89.29 & 0.57 & 0.0377 & 1.25 \\
gru & dowjones & 12\_months & 0.89 & 126.84 & 91.94 & 0.58 & 0.0347 & -0.15 \\
gru & no-sentiment & 1\_months & 0.84 & 150.85 & 109.49 & 0.55 & 0.1213 & -0.54 \\
gru & no-sentiment & 3\_months & 0.83 & 157.38 & 111.45 & 0.54 & 0.2588 & -0.44 \\
gru & no-sentiment & 6\_months & 0.85 & 147.14 & 112.09 & 0.50 & 0.9435 & 0.08 \\
gru & no-sentiment & 12\_months & 0.82 & 161.15 & 125.29 & 0.55 & 0.1693 & -0.52 \\
bilstm & chinaservice & 1\_months & 0.89 & 125.11 & 89.85 & 0.57 & 0.0478 & 0.50 \\
bilstm & chinaservice & 3\_months & 0.88 & 130.69 & 95.01 & 0.57 & 0.0328 & 0.07 \\
bilstm & chinaservice & 6\_months & 0.89 & 126.80 & 96.00 & 0.57 & 0.0534 & 0.33 \\
bilstm & chinaservice & 12\_months & 0.88 & 130.00 & 98.25 & 0.54 & 0.2193 & 0.56 \\
bilstm & reuters & 1\_months & 0.89 & 126.79 & 91.59 & 0.55 & 0.1213 & 0.70 \\
bilstm & reuters & 3\_months & 0.88 & 132.66 & 92.02 & 0.60 & 0.0041 & 2.32 \\
bilstm & reuters & 6\_months & 0.89 & 126.92 & 88.74 & 0.55 & 0.1344 & -0.46 \\
bilstm & reuters & 12\_months & 0.89 & 128.64 & 95.47 & 0.52 & 0.6141 & -0.67 \\
bilstm & dowjones & 1\_months & 0.88 & 130.12 & 93.37 & 0.52 & 0.5751 & 0.19 \\
bilstm & dowjones & 3\_months & 0.88 & 129.93 & 92.71 & 0.54 & 0.2035 & -0.34 \\
bilstm & dowjones & 6\_months & 0.89 & 126.94 & 88.22 & 0.58 & 0.0261 & -0.23 \\
bilstm & dowjones & 12\_months & 0.89 & 123.87 & 91.57 & 0.54 & 0.2193 & -0.21 \\
bilstm & no-sentiment & 1\_months & 0.89 & 124.68 & 90.66 & 0.55 & 0.1213 & -0.01 \\
bilstm & no-sentiment & 3\_months & 0.88 & 131.74 & 89.92 & 0.57 & 0.0328 & 0.05 \\
bilstm & no-sentiment & 6\_months & 0.89 & 125.87 & 89.84 & 0.55 & 0.1760 & 0.75 \\
bilstm & no-sentiment & 12\_months & 0.89 & 127.39 & 91.64 & 0.53 & 0.4277 & 0.02 \\
convlstm & chinaservice & 1\_months & 0.90 & 117.84 & 82.64 & 0.58 & 0.0158 & -0.29 \\
convlstm & chinaservice & 3\_months & 0.89 & 124.18 & 87.29 & 0.57 & 0.0328 & -0.11 \\
convlstm & chinaservice & 6\_months & 0.87 & 134.19 & 97.63 & 0.56 & 0.1007 & 0.39 \\
convlstm & chinaservice & 12\_months & 0.89 & 125.63 & 93.95 & 0.54 & 0.2788 & 0.35 \\
convlstm & reuters & 1\_months & 0.90 & 117.84 & 82.64 & 0.58 & 0.0158 & -0.29 \\
convlstm & reuters & 3\_months & 0.88 & 132.64 & 90.43 & 0.57 & 0.0467 & 0.06 \\
convlstm & reuters & 6\_months & 0.87 & 134.80 & 96.54 & 0.56 & 0.0740 & -0.51 \\
convlstm & reuters & 12\_months & 0.88 & 132.48 & 97.26 & 0.55 & 0.1283 & -0.20 \\
convlstm & dowjones & 1\_months & 0.90 & 117.84 & 82.64 & 0.58 & 0.0158 & -0.29 \\
convlstm & dowjones & 3\_months & 0.88 & 132.64 & 90.43 & 0.57 & 0.0467 & 0.06 \\
convlstm & dowjones & 6\_months & 0.87 & 134.80 & 96.54 & 0.56 & 0.0740 & -0.51 \\
convlstm & dowjones & 12\_months & 0.88 & 134.74 & 97.52 & 0.56 & 0.0954 & -0.59 \\
convlstm & no-sentiment & 1\_months & 0.90 & 117.59 & 82.30 & 0.59 & 0.0104 & -0.29 \\
convlstm & no-sentiment & 3\_months & 0.88 & 129.37 & 89.36 & 0.56 & 0.0651 & 0.22 \\
convlstm & no-sentiment & 6\_months & 0.88 & 132.81 & 96.95 & 0.55 & 0.1344 & 0.24 \\
convlstm & no-sentiment & 12\_months & 0.88 & 130.05 & 96.46 & 0.52 & 0.6141 & 1.02 \\
tft & chinaservice & 1\_months & 0.88 & 132.26 & 98.11 & 0.57 & 0.0337 & 0.05 \\
tft & chinaservice & 3\_months & 0.87 & 133.48 & 96.20 & 0.58 & 0.0226 & -0.39 \\
tft & chinaservice & 6\_months & 0.88 & 130.66 & 96.20 & 0.55 & 0.1760 & -0.58 \\
tft & chinaservice & 12\_months & 0.88 & 132.34 & 96.14 & 0.57 & 0.0497 & -0.41 \\
tft & reuters & 1\_months & 0.88 & 131.63 & 95.42 & 0.56 & 0.0906 & -0.19 \\
tft & reuters & 3\_months & 0.88 & 128.74 & 90.06 & 0.57 & 0.0328 & -0.59 \\
tft & reuters & 6\_months & 0.88 & 130.66 & 92.73 & 0.54 & 0.2265 & 0.00 \\
tft & reuters & 12\_months & 0.88 & 131.93 & 93.59 & 0.58 & 0.0347 & -0.35 \\
tft & dowjones & 1\_months & 0.87 & 134.40 & 95.62 & 0.54 & 0.2058 & -0.23 \\
tft & dowjones & 3\_months & 0.88 & 132.15 & 96.22 & 0.55 & 0.1573 & -0.44 \\
tft & dowjones & 6\_months & 0.87 & 134.51 & 98.03 & 0.53 & 0.4348 & -0.55 \\
tft & dowjones & 12\_months & 0.88 & 134.85 & 96.60 & 0.54 & 0.2193 & -0.73 \\
tft & no-sentiment & 1\_months & 0.88 & 131.48 & 94.61 & 0.54 & 0.2611 & -0.58 \\
tft & no-sentiment & 3\_months & 0.88 & 132.73 & 93.59 & 0.57 & 0.0467 & -0.82 \\
tft & no-sentiment & 6\_months & 0.88 & 130.88 & 96.12 & 0.53 & 0.3557 & -0.53 \\
tft & no-sentiment & 12\_months & 0.88 & 133.36 & 96.45 & 0.55 & 0.1693 & -0.52 \\
\bottomrule
\addlinespace[2ex]
\caption{Trading simulation results for the best hyperparameter combination for each model, sentiment source and time window. The p-value here measures the statistical significance of the model's directional accuracy relative to random chance (50\%). The TotRet metric shows the total return of the trading strategy over the backtest period.}
\label{tab:full_trading_results}
\end{longtable}
}

\twocolumn

\end{document}